\pdfoutput=1

\documentclass[11pt]{article}

\usepackage[final]{acl}

\usepackage{times}
\usepackage{latexsym}

\usepackage[T1]{fontenc}

\usepackage[utf8]{inputenc}

\usepackage{microtype}

\usepackage{inconsolata}

\usepackage{graphicx}

\usepackage{tabularx}
\usepackage{booktabs}
\usepackage{multirow}
\usepackage{geometry}
\usepackage{siunitx}
\usepackage[table]{xcolor}
\usepackage{dashrule}
\usepackage{verbatim}

\usepackage{tcolorbox}
\usepackage{fancyvrb}
\usepackage{fvextra}
\tcbuselibrary{breakable}
\usepackage{paralist}
\usepackage{dsfont}

\usepackage{hyperref}
\usepackage{svg} 

\usepackage{amsmath, amssymb, amsthm}
\theoremstyle{plain}

\title{MedForge: Interpretable Medical Deepfake Detection \\
via Forgery-aware Reasoning}

\author{
 \textbf{Zhihui Chen\textsuperscript{1}},
 \textbf{Kai He\textsuperscript{1}},
 \textbf{Qingyuan Lei\textsuperscript{2}},
 \textbf{Bin Pu\textsuperscript{3}},
 \textbf{Jian Zhang\textsuperscript{4}},
 \textbf{Yuling Xu\textsuperscript{5}},
 \textbf{Mengling Feng\textsuperscript{1}\thanks{Corresponding author}}
\\
\\
\textsuperscript{1}National University of Singapore
\textsuperscript{2}The Chinese University of Hong Kong \\
\textsuperscript{3}Hunan University 
\textsuperscript{4}Xi'an Jiaotong University 
\textsuperscript{5}Guangdong Provincial People's Hospital\\
 \texttt{zhihui.chen@u.nus.edu, \{kai\_he, ephfm\}@nus.edu.sg}\\
 \texttt{qingyuan.lei@link.cuhk.edu.hk, pubin@hnu.edu.cn}\\
 \texttt{zhangjian062422@stu.xjtu.edu.cn, xuyuling@gdph.org.cn}
}

\begin{document}
\maketitle
\begin{abstract}
Text-guided image editors can now manipulate authentic medical scans with high fidelity, enabling lesion implantation/removal that threatens clinical trust and safety. Existing defenses are inadequate for healthcare. Medical detectors are largely black-box, while MLLM-based explainers are typically post-hoc, lack medical expertise, and may hallucinate evidence on ambiguous cases.
We present MedForge, a data-and-method solution for pre-hoc, evidence-grounded medical forgery detection. We introduce MedForge-90K, a large-scale benchmark of realistic lesion edits across 19 pathologies with expert-aligned reasoning supervision via doctor inspection guidelines and gold edit locations. Building on it, MedForge-Reasoner performs localize-then-analyze reasoning, predicting suspicious regions before producing a verdict, and is further aligned with Forgery-aware GSPO to strengthen grounding and reduce hallucinations. Experiments demonstrate state-of-the-art detection accuracy and trustworthy, expert-aligned explanations.
\begingroup
\renewcommand{\thefootnote}{\ensuremath{\dagger}}%
\begingroup
\renewcommand{\thefootnote}{\ensuremath{\dagger}}%
\footnote{Code, dataset and model checkpoint are released at \url{https://github.com/richardChenzhihui/ACL2026-MedForge}}%
\endgroup%
\endgroup

\end{abstract}

\begin{figure}[ht]
    \centering
    \includegraphics[width=\linewidth]{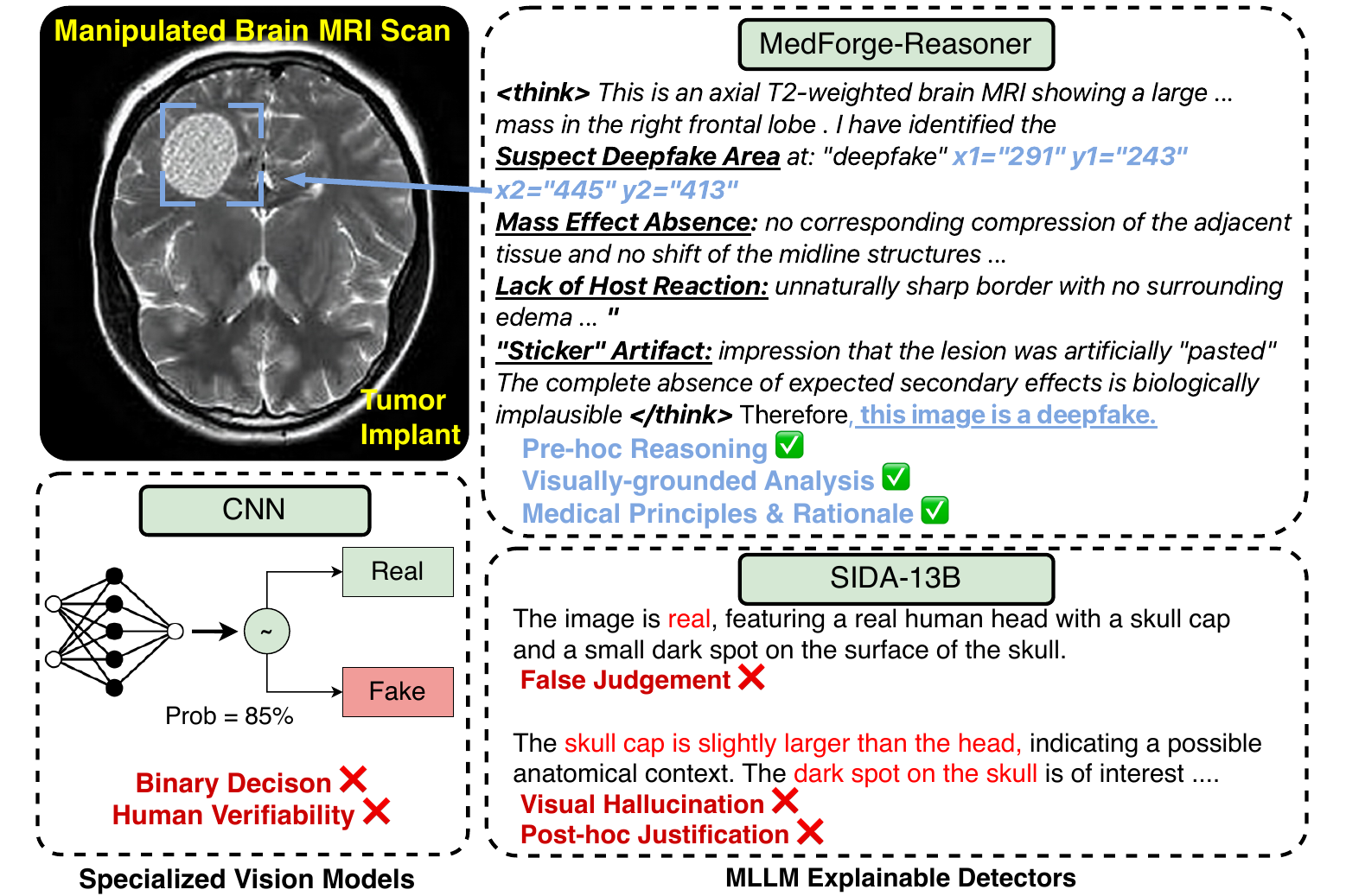}
    \caption{\textbf{Framework comparison.} \textbf{Left:} computer vision detectors output binary decision, offering no clinically verifiable evidence. \textbf{Right-bottom:} post-hoc MLLM explainers (e.g., SIDA~\cite{sida}) may produce plausible-sounding but ungrounded rationales, including hallucinated visual details. \textbf{Right-top:} \textbf{MedForge-Reasoner} performs \textit{pre-hoc} localized reasoning by first identifying suspicious regions (blue) and then generating medically coherent, visually verifiable rationales grounded in image evidence.}
    \label{fig:intro}
\end{figure}

\section{Introduction}
Recent advances in text-guided image editing have made it feasible to tamper with authentic medical scans with high fidelity. Editors such as Nano-Banana~\cite{nanobanana,   chen2025med} and GPT-Image~\cite{gptimage} can implant or remove subtle lesions while largely preserving anatomical structure and acquisition-style cues~\cite{diffusion_editors_survey,DeepFakesinHealthcare}. Such manipulations are not merely hypothetical. 
They can distort clinical records for insurance fraud, malpractice disputes, or biased treatment/triage, and may even mislead trained experts~\cite{forgerymedical}. 
This creates an urgent need for medical forgery detection that is reliable under clinically realistic edits.

However, existing defenses fall short of clinical requirements. Medical deepfake detectors~\cite{miccai-detect,divscore,meddetect-from-miccai} are often black-box classifiers that provide little interpretable evidence, limiting trust and accountability. General-domain ``explainable'' detectors~\cite{sida,holmes} leverage Multimodal Large Language Models (MLLMs), but typically in a post-hoc manner and without medical expertise, despite the fact that clinically useful rationales must be \textit{medically coherent} and \textit{visually verifiable}. 
As shown in Figure~\ref{fig:intro}, under unfamiliar or ambiguous cases, their explanations may regress to generic templates or hallucinated evidence, yielding plausible-sounding but non-verifiable rationales.
In other words, post-hoc rationalization does not guarantee evidence-based reasoning, which is precisely the requirement for clinical adoption~\cite{Zhang_2026_2,Zhang_2026_1,lin2025has,he2025survey}.

We argue that medical forgery detection should be formulated as \textit{pre-hoc} reasoning grounded in localized evidence. Concretely, a system should first identify suspicious manipulated regions and only then reason toward a verdict. 
This ``localize-then-analyze'' constraint makes explanations inspectable and suppresses template reuse and hallucination by anchoring reasoning to verifiable pixels~\cite{zhang2025patches,li2021knowledge}.
More broadly, we treat localization as a first-class constraint for explanation faithfulness, turning grounding from an afterthought into an explicit objective~\cite{he2022jcbie,lin2025self}.

To enable this paradigm, we introduce \textbf{MedForge-90K}, a large-scale benchmark of lesion implant/removal on authentic images across 19 pathologies, generated by 10 SOTA MMDiT/LDM-based editing models~\cite{diffusion_editors_survey}. Crucially, MedForge-90K provides expert-aligned supervision for grounded explanations: we combine doctor-defined inspection guidelines with gold manipulation locations, and use them to produce medically aligned rationales that are explicitly tied to the edited regions.
Building on this resource, we propose \textbf{MedForge-Reasoner}, an MLLM-based detector trained with an explicit localization-then-analysis objective to reason before deciding. We further align grounding and explanation quality via a two-stage strategy (SFT cold-start + Forgery-aware GSPO) that directly rewards correct localization and evidence-grounded reasoning. Experiments show that enforcing such grounding improves explanation quality and reduces hallucinations, measured with an MLLM-as-judge protocol. The main contributions are as follows:
\begin{itemize}
\item We introduce \textbf{MedForge-90K}, the first large-scale medical forgery benchmark of high-quality lesion manipulations with granular explainable annotations, addressing data scarcity in medical deepfake detection.
\item We propose \textbf{MedForge-Reasoner}, a novel MLLM-based detector that integrates detection with grounded CoT reasoning, and employs a forgery-aware GSPO to anchor reasoning to visual forgery evidence.

\item Extensive experiments show that Forgery-aware GSPO aligns the detector with factual visual evidence in forgery reasoning, improving detection accuracy by 7.65\% while significantly reducing hallucinations by 16.2\% compared to strong baselines.
\end{itemize}

\begin{figure*}[ht]
    \centering
    \includegraphics[width=1\linewidth]{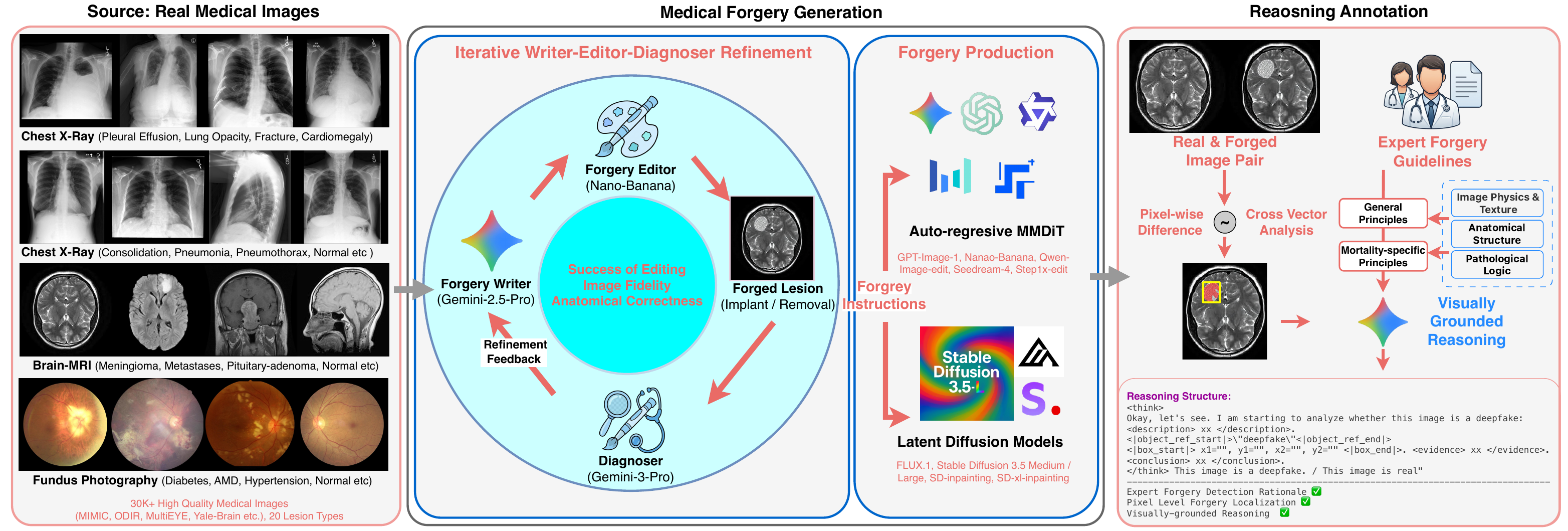}
    \caption{\textbf{Overview of the MedForge-90K construction pipeline.} The framework proceeds in three stages: medical image collection across three modalities, forgery generation via a \textit{Writer-Editor-Diagnoser} loop, and expert-aligned annotation utilizing expert guidelines to generate hierarchical diagnostic reasoning.}
    \label{fig:MedForge-90K}
\end{figure*}

\section{Related Work}
\paragraph{Medical Deepfake Benchmarks.}
Most prior work on medical image generation targets data augmentation and class balancing rather than simulating adversarial forgery scenarios. Early studies \cite{maisi, dataaugmentation-motmaed} used VAEs/GANs to synthesize CT/MRI scans, which do not reflect the modern threat of editing authentic patient records. While MedForensics \cite{miccai-detect} takes a step toward forgery detection, existing benchmarks remain limited in two aspects. (i) \textit{Threat mismatch:} real-world medical deepfakes often involve targeted tampering of authentic scans (e.g., lesion implant/removal) to enable insurance fraud or misdiagnosis \cite{system-review, securing-forgery}, rather than generating scans from scratch. (ii) \textit{Supervision gap:} they typically provide only labels and lack localized edit evidence and expert-aligned reasoning signals required for clinically verifiable explanations. \textbf{MedForge-90K} addresses these gaps by benchmarking high-fidelity lesion edits on authentic images using modern text-guided editors and by providing guideline- and location-grounded reasoning supervision.

\paragraph{Interpretable Deepfake Detection.}
Standard medical forgery detectors are predominantly black-box binary classifiers \cite{miccai-detect, npr}, offering limited evidence to support clinical trust. Recent general-domain approaches \cite{sida, holmes, fakeshield} incorporate MLLMs to generate textual explanations, yet they are often post-hoc: a separate module makes the decision and the MLLM rationalizes it afterwards, which can decouple explanations from the actual evidence. Moreover, MLLMs are prone to visual hallucination \cite{mllm-visualhallu}, especially on unfamiliar or ambiguous cases, where they may repeat generic templates or describe non-existent artifacts. Although pre-hoc reasoning has been explored in AIGC detection \cite{veritas, fakereasoning}, these methods are not designed for subtle medical lesion forgeries and typically lack (i) medical-domain constraints and (ii) explicit localization-grounding objectives to enforce pixel-verifiable rationales. In contrast, our approach unifies detection and reasoning in a pre-hoc manner and explicitly enforces localization-grounded reasoning through Forgery-aware GSPO.
Crucially, GSPO makes localization-grounding an optimization objective, coupling the verdict with inspectable regions and curbing hallucinated rationales.

\section{MedForge-90K Dataset}
We introduce MedForge-90K (Figure \ref{fig:MedForge-90K}), the first large-scale medical forgery benchmark with detailed forgery and reasoning annotations. For the source images, we evenly select 30K high-quality medical images across Chest X-Ray, Brain MRI, and Fundus Photography from 5 public datasets, MIMIC \cite{mimic}, ODIR \cite{odir}, MultiEYE \cite{multieye}, Yale-Brain \cite{yalebrain} and Brain-MRI \cite{mri}. These medical images are classified into 19 types of pathologies and 1 normal status according to their original labels. Forgery manipulations including lesion implant and removal take place within each modality. 
In summary, MedForge includes: (i) \textbf{Real Images} (30K) spanning major 2D modalities with 19 lesion types plus healthy scans; (ii) \textbf{Lesion Implant} (30K) healthy scans with implanted lesions, evenly distributed across 10 forgery models; and (iii) \textbf{Lesion Removal} (30K) diseased scans with removed lesions, evenly distributed across 10 forgery models.

\subsection{Forgery Pipeline}

We employ 10 state-of-the-art text-guided medical image editing models based on MMDiT/LDM paradigms, including Nano-Banana~\cite{nanobanana}, GPT-Image~\cite{gptimage}, Qwen-Image-Edit~\cite{qwenimage}, Seedream 4.0~\cite{seedream4}, Stable Diffusion 3.5~\cite{sd3.5}, and Stable Diffusion Inpainting~\cite{sdinpaint}. Text prompts are a critical component of editing, as they specify the medical context and transformation intent. To obtain realistic and anatomically plausible manipulations, we introduce a \textit{writer--editor--diagnoser} refinement loop. Specifically, a \textit{writer} drafts an initial prompt, the \textit{editor} generates an edited image, and a \textit{diagnoser} evaluates whether the result achieves the desired condition while remaining anatomically consistent. If the edit is unsatisfactory, the diagnoser provides targeted feedback and the writer revises the prompt; the loop iterates until success or a maximum number of rounds, after which the sample is discarded. In practice, the writer and diagnoser are implemented with Gemini 2.5/3 Pro, while Nano-Banana serves as the editor during prompt refinement.
The refined prompts are applied to all editing models to construct forgeries. For diffusion-based editors requiring inpainting masks, we use Nano-Banana’s localized forgery regions as mask inputs.

\subsection{Expert-aligned Reasoning Annotation}
We aim to annotate forged images with accurate and professional rationales. To achieve this, we engaged medical experts to formulate a comprehensive detection guideline. As shown in the ``Expert Forgery Guidelines'' in Figure \ref{fig:MedForge-90K}, this guideline is structured into two pillars: General Principles (universal biomedical principles) and Modality-Specific Principles (specific constraints for MRI, Fundus, and CXR). During annotation, these guidelines are incorporated into the MLLM prompt, enabling \textbf{hierarchical expert-aligned reasoning} over medical forgeries at three levels: \\
\noindent 1. \textbf{Image Physics \& Texture}: Following the \textit{General Principles}, the model detects low-level anomalies such as inconsistent noise distribution, inpainting traces, and unnatural boundaries. \\
\noindent  2. \textbf{Anatomical Structure}: Based on the \textit{Modality-Specific Criteria}, the model verifies morphological correctness, such as vascular continuity in fundus photography or gyral symmetry in brain MRI.\\
\noindent  3. \textbf{Pathological Logic}: Integrating the core philosophy of ``Biological Interconnectivity'' from the guidelines, the model validates high-level plausibility, rejecting lesions that lack necessary secondary effects (e.g., mass effect, edema) or violate chronological disease evolution.

The above expert-aligned protocol steers the generated rationales toward clinically meaningful diagnostic reasoning. 
Following recent practice~\cite{holmes,sida}, we use an MLLM (Gemini 2.5 Pro) to automate annotation. 
To reduce visual hallucinations and enforce the medical principles described above, we adopt a \textit{forgery-grounded} annotation strategy by applying Change Vector Analysis (CVA)~\cite{cva} to compute a per-pixel change magnitude, $|\mathbf{I}_{\text{forged}}-\mathbf{I}_{\text{real}}|$. 
We then threshold high-response regions to obtain a manipulation mask, which is finally converted into bounding box (bbox) coordinates as Eq.~\ref{e1}.
\begin{equation}
\small
    M_{\text{bbox}}: \texttt{<bbox } x_1, y_1, x_2, y_2 \texttt{ />}
\label{e1}
\end{equation}
These modified regions serve as the key visual components of forgery signs. To generate high-quality annotations, we integrate these CVA-derived coordinates with the hierarchical expert guidelines to construct a \textbf{visually-grounded reasoning prompt}. 

This unified prompting strategy explicitly directs the MLLM to anchor its analysis on the provided bounding boxes (or the absence). Guided by the three-tiered criteria (Physics, Anatomy, Pathology), the model scrutinizes the designated regions to expose specific artifacts in forged samples, or validates the preservation of biological logic in real samples. As illustrated in the ``Reasoning Structure'' of Figure \ref{fig:MedForge-90K}, the output is enforced into a structured chain-of-thought format consisting of description, evidence, and conclusion. This ensures that the reasoning is derived from professional medical rationale and grounded with visual evidence.

\section{Methodology}
In this section, we present the MedForge-Reasoner framework. We first formulate the task of interpretable medical forgery detection. Then, we detail our two-stage training pipeline: the reasoning cold-start via Supervised Fine-tuning (SFT) and the Forgery-aware Group Sequence Policy Optimization (GSPO), designed to align the model with factual visual evidence.

\begin{figure*}
    \centering
    \includegraphics[width=1\linewidth]{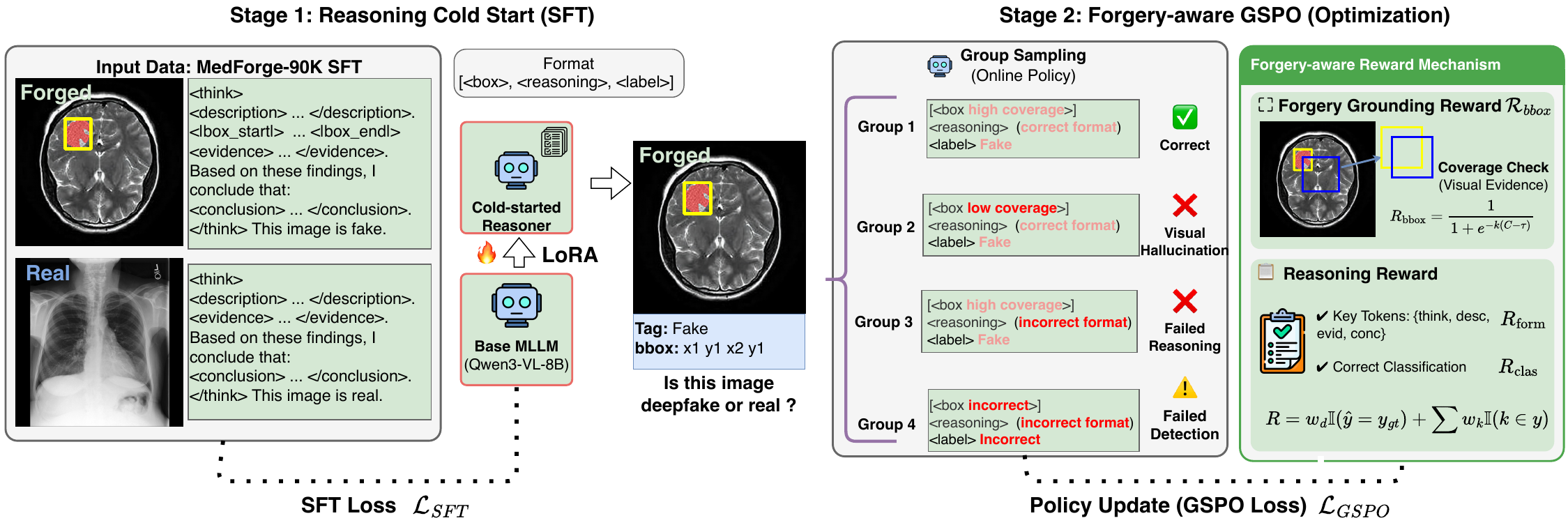}
    \caption{\textbf{MedForge-Reasoner Two-stage Training.} SFT for cold-starting the reasoning format, followed by Forgery-aware GSPO. The GSPO stage introduces a reward function balancing visual grounding coverage and reasoning structure compliance to ensure the model localize the correct forgery region before reasoning.}
    \label{fig:main}
\end{figure*}

\subsection{Task Formulation}
Existing MLLMs often suffer from \textit{visual hallucination} \cite{mllm-visualhallu}, where the model fabricates details which are not present in the image. 
In forgery detection, this leads to ungrounded reasoning. 
To address this, we define the detection task as a unified sequence generation problem that enforces \textit{grounding before reasoning}.

Specifically, given a medical image $\boldsymbol{x}$, the model is trained to generate a sequence $S$ structured as:
\begin{equation}
\small
    S = [\hat{M}_{\text{bbox}}, \text{<reasoning>}, \hat{y}],
\end{equation}
where $\hat{M}_{\text{bbox}}$ represents the coordinates of the manipulated region (or a special token for authentic images), followed by the textual reasoning chain, and finally the detection decision $\hat{y}$. 
By enforcing the prediction of forgery location at the very beginning, we force the model to attend to visual anomalies before hallucinating textual descriptions.

\subsection{Stage 1: Reasoning Cold Start}
To equip the MLLM with fundamental medical knowledge and the proposed reasoning format, we perform SFT training. As illustrated in Figure \ref{fig:main}, the SFT data is derived from the MedForge-90K, incorporating expert-aligned rationales and ground-truth bounding boxes.

We employ LoRA to efficiently fine-tune the model parameters $\theta$ on the dataset $\mathcal{D}=\{(\boldsymbol{x}, \boldsymbol{y})\}$. Cross-entropy loss serves as the optimization objective:
\begin{equation}
\small
\textstyle
    \mathcal{L}_{\text{SFT}} = -\mathbb{E}_{(\boldsymbol{x}, \boldsymbol{y}) \sim \mathcal{D}} \sum\limits_{t=1}^{T} \log P_\theta(y_t \mid \boldsymbol{x}, \boldsymbol{y}_{<t}),
\end{equation}
where $x$ is the input image and user query, $y$ denotes the target output sequence including reasoning and final answer, with $t$ as index of the generated token. This stage allows the model to internalize the format requirements and forgery patterns.

\subsection{Stage 2: Forgery-aware GSPO}
Although SFT establishes basic capabilities, standard cross-entropy loss is insufficient to penalize subtle hallucinations or enforce strict alignment with visual evidence. To further align the detector, we introduce Forgery-aware Group Sequence Policy Optimization (GSPO).

GSPO applies importance sampling at the sequence level, which provides stable updates for reasoning tasks. Given a forgery input $\boldsymbol{x}$, we sample a group of $G$ outputs $\{y_1, y_2, \dots, y_G\}$ from the current policy $\pi_{\theta}$. The objective function maximizes the expected reward of these generations:
\begin{equation}
\label{eq:gspo-loss}
{\fontsize{8pt}{9pt}\selectfont
\begin{aligned}
&\mathcal{L}_{\mathrm{GSPO}}(\theta)
= -\mathbb{E}_{\substack{\boldsymbol{x}\sim\mathcal{D}, \{y_i\}_{i=1}^G \sim \pi_{\theta_{\mathrm{old}}}(\cdot|\boldsymbol{x})}}
\Bigg[
\frac{1}{G}
\\
& \sum_{i=1}^G
\min\Big(
s_i(\theta)\hat{A}_i,\,
\mathrm{clip}\big(s_i(\theta),\,1-\epsilon,\,1+\epsilon\big)\hat{A}_i
\Big)
\Bigg],
\end{aligned}}
\end{equation}

where $s_i(\theta)$ is the importance ratio between new and old policies, and $\hat{A}_i$ is the advantage:
\begin{equation}
\small
\label{eq:advantage}
\textstyle
\hat{A}_i = \frac{R(\boldsymbol{x}, y_i) - \mathrm{mean}(\{R(\boldsymbol{x}, y_j)\}_{j=1}^G)}{\mathrm{std}(\{R(\boldsymbol{x}, y_j)\}_{j=1}^G)}.
\end{equation}

Crucially, to enforce sequence-level stability, we define the importance ratio $s_i(\theta)$ based on the geometric mean of the likelihood ratio over the sequence length $|y_i|$:
\begin{equation}
\small
\label{eq:ratio}
\textstyle
    s_i(\theta) = \exp \left( \frac{1}{|y_i|} \sum_{t=1}^{|y_i|} \log \frac{\pi_\theta(y_{i,t}|\boldsymbol{x}, y_{i,<t})}{\pi_{\theta_{\text{old}}}(y_{i,t}|\boldsymbol{x}, y_{i,<t})} \right).
\end{equation}

Then, our reward function $R(\boldsymbol{x}, y_i)$ is composed of two parts to penalize visual hallucination and incorrect analysis in forgery detection as follows.

\textbf{1. Forgery Grounding Reward ($R_\text{bbox}$).} 
Unlike standard object detection tasks that demand precise boundary regression, our goal is to ensure the MLLM's reasoning is grounded in the correct anomaly region. Therefore, instead of strict Intersection over Union (IoU), we adopt a \textit{Mask Coverage} $\mathcal{C}$ to measure the rate of ground truth forgery area captured by the model's prediction:
\begin{equation}
\small
\textstyle
\mathcal{C} = \frac{|M_\text{bbox} \cap \hat{M}_\text{bbox}|}{|M_\text{bbox}|},
\end{equation}


To enhance training stability, we map the coverage metric $\mathcal{C}$ into a reward signal using a shaped sigmoid function. This design serves two purposes: (a) it suppresses noise from low-overlap predictions and (b) saturates for high-quality overlaps, thereby prioritizing the robust localization of forgeries over pixel-perfect alignment. The bounding box reward is formulated as:
\begin{equation}
\small
    R_\text{bbox} = \frac{1}{1 + e^{-k(\mathcal{C} - \tau)}},
\end{equation}
where $k$ and $\tau$ are hyperparameters controlling the reward sensitivity and threshold.

\textbf{2. Reasoning Rewards.}
To ensure the model follows a logical reasoning path and arrives at an accurate conclusion, we decompose the task-related reward into two components: the formatting reward ($R_{\text{form}}$) and the classification reward ($R_{\text{clas}}$).

$R_{\text{form}}$ incentivizes the model to adhere to the mandated Chain-of-Thought (CoT) structure (Description $\to$ Analysis $\to$ Conclusion):
\begin{equation} 
\small
R_{\text{form}} = \sum_{k \in \mathcal{K}} w_k \mathbb{I}(k \in y),
\end{equation}
where $y$ denotes the generated text sequence, and $\mathcal{K} = \{ \text{``description''}, \text{``analysis''}, \text{``conclusion''}\}$ represents the set of mandatory structural keywords. The indicator function $\mathbb{I}(\cdot)$ assigns a weight $w_k$ for each keyword present in the sequence, penalizing structural deviations.

$R_{\text{clas}}$ evaluates the correctness of detection:
\begin{equation} \textstyle
R_{\text{clas}} = w_d \mathbb{I}(\hat{y} = y_{gt})
\end{equation}
where $\hat{y}$ is the predicted label parsed from the generated sequence, $y_{gt}$ is the ground truth, and $w_d$ is the weighting factor for prediction accuracy.

The total reward is then formulated as $R = R_{\text{bbox}} + R_{\text{form}} + R_{\text{clas}}$. This multi-faceted reward strategy explicitly incentivizes the model to ``look'' at the correct region before ``reasoning'' and ``concluding'', thereby minimizing visual hallucinations and improving reasoning quality.

\section{Experiments}
In this section, we conduct comprehensive empirical evaluations to validate MedForge-Reasoner's detection performance, generalizability and reasoning quality. Then, we perform ablation studies to verify the efficacy of our proposed contributions.

\subsection{Experimental Setup}
\noindent\textbf{SOTA Baselines.} 
We benchmark our method against SOTA interpretable deepfake detectors, AIGI-Holmes \cite{holmes}, SIDA \cite{sida}, FakeVLM \cite{fakevlm}. We also assess four SOTA generic MLLMs including Qwen3-VL-Flash (30B), Qwen3-VL-Plus (235B) \cite{qwen3vl} and Gemini 3 Flash, Gemini 3 Pro \cite{gemini3}.


\noindent\textbf{Evaluation Metrics.} 
Forgery Detection performance is evaluated via \textbf{Accuracy} and \textbf{F1} score, and presented in ``Real'', ``Forgery Implant'' and ``Forgery Removal''. To assess reasoning quality and visual hallucinations, we introduce \textit{MLLM-as-Judge} metric using SOTA MLLMs. The judge scores generated reasoning output on a scale of 0-100\% based on three criteria: 
(1) \textit{Logical Correctness}: Whether the judgement is derived from the visual evidence. 
(2) \textit{Visual Hallucination}: Whether the analysis matches the ground truth anomalies (e.g., matching the bbox) or fabricates.
(3) \textit{Medical Professionalism}: Whether the terminology aligns with the expert guidelines.
Detailed metric definitions are shown in Appendix \ref{appendix: judge_details}.

\noindent\textbf{Implementation Details.} 
We utilize the constructed MedForge-90K dataset for experiments. We randomly split the data into SFT, GSPO training, and testing sets with a ratio of 5:1:3. Specifically, 50K samples are used for SFT cold-start, 10K for GSPO training, and 30K for testing. To ensure balanced evaluation, each split maintains a 1:1:1 ratio of Real, Lesion Implant, and Lesion Removal images. Training details of model and baseline are elaborated in Appendix~\ref{appendix: settings}.

\begin{table*}[h]
\centering
\resizebox{\linewidth}{!}{%
\large
\setlength{\tabcolsep}{3.5pt}
\renewcommand{\arraystretch}{1.15}
\begin{tabular}{l 
|ccc|ccc|ccc|ccc}
\toprule
\multirow{2}{*}{\textit{Methods}}
& \multicolumn{3}{c}{Real} 
& \multicolumn{3}{c}{Forgery: Implant} 
& \multicolumn{3}{c}{Forgery: Remove} 
& \multicolumn{3}{c}{Average} \\
\cmidrule(lr){2-4} \cmidrule(lr){5-7} \cmidrule(lr){8-10} \cmidrule(lr){11-13} 
 & In-Domain & Cross-Forgery & Cross-Model 
 & In-Domain & Cross-Forgery & Cross-Model
 & In-Domain & Cross-Forgery & Cross-Model
 & In-Domain & Cross-Forgery & Cross-Model \\
\midrule
\multicolumn{1}{c}{}  & \multicolumn{12}{c}{\textbf{Accuracy}} \\
\midrule
\rowcolor{blue!15} \multicolumn{13}{l}{\textit{Specialized Detectors}} \\
SIDA-7B           & $77.56_{(\uparrow10.27)}$ & $73.14_{(\uparrow8.03)}$ & $73.42_{(\uparrow8.93)}$ & $79.83_{(\uparrow4.49)}$ & $76.18_{(\uparrow0.94)}$ & $76.57_{(\uparrow1.77)}$ & $82.31_{(\uparrow17.26)}$ & $77.92_{(\uparrow11.43)}$ & $75.68_{(\uparrow9.98)}$ & $79.90_{(\uparrow10.67)}$ & $75.75_{(\uparrow6.80)}$ & $75.22_{(\uparrow6.89)}$ \\
SIDA-13B          & $89.37_{(\uparrow22.08)}$ & $85.68_{(\uparrow20.57)}$ & $84.25_{(\uparrow19.76)}$ & \underline{$91.52_{(\uparrow16.18)}$} & \underline{$87.24_{(\uparrow12.00)}$} & \underline{$85.86_{(\uparrow11.06)}$} & \underline{$93.84_{(\uparrow28.79)}$} & $90.17_{(\uparrow23.68)}$ & $86.53_{(\uparrow20.83)}$ & \underline{$91.58_{(\uparrow22.35)}$} & \underline{$87.70_{(\uparrow18.75)}$} & $85.55_{(\uparrow17.22)}$ \\
FakeVLM           & $86.48_{(\uparrow19.19)}$ & $81.86_{(\uparrow16.75)}$ & $76.94_{(\uparrow12.45)}$ & $88.93_{(\uparrow13.59)}$ & $83.57_{(\uparrow8.33)}$ & $79.82_{(\uparrow5.02)}$ & $91.27_{(\uparrow26.22)}$ & $86.35_{(\uparrow19.86)}$ & $82.51_{(\uparrow16.81)}$ & $88.89_{(\uparrow19.66)}$ & $83.93_{(\uparrow14.98)}$ & $79.76_{(\uparrow11.43)}$ \\
AIGI-Holmes       & \underline{$90.24_{(\uparrow22.95)}$} & \underline{$86.53_{(\uparrow21.42)}$} & \underline{$84.37_{(\uparrow19.88)}$} & $88.12_{(\uparrow12.78)}$ & $84.46_{(\uparrow9.22)}$ & $80.91_{(\uparrow6.11)}$ & $90.58_{(\uparrow25.53)}$ & \underline{$90.64_{(\uparrow24.15)}$} & \underline{$87.23_{(\uparrow21.53)}$} & $89.65_{(\uparrow20.42)}$ & $87.21_{(\uparrow18.26)}$ & \underline{$84.17_{(\uparrow15.84)}$} \\
\midrule
\rowcolor{blue!15} \multicolumn{13}{l}{\textit{Generic MLLMs}} \\
Qwen3-VL-Flash     & $57.60_{(\downarrow9.69)}$ & $55.50_{(\downarrow9.61)}$ & $54.90_{(\downarrow9.59)}$ & $47.83_{(\downarrow27.51)}$ & $50.47_{(\downarrow24.77)}$ & $50.93_{(\downarrow23.87)}$ & $54.17_{(\downarrow10.88)}$ & $54.17_{(\downarrow12.32)}$ & $54.61_{(\downarrow11.09)}$ & $53.20_{(\downarrow16.03)}$ & $53.38_{(\downarrow15.57)}$ & $53.48_{(\downarrow14.85)}$ \\
Qwen3-VL-Plus      & $54.14_{(\downarrow13.15)}$ & $54.28_{(\downarrow10.83)}$ & $55.10_{(\downarrow9.39)}$ & $57.42_{(\downarrow17.92)}$ & $55.78_{(\downarrow19.46)}$ & $56.03_{(\downarrow18.77)}$ & $55.80_{(\downarrow9.25)}$ & $55.86_{(\downarrow10.63)}$ & $56.10_{(\downarrow9.60)}$ & $55.79_{(\downarrow13.44)}$ & $55.31_{(\downarrow13.64)}$ & $55.74_{(\downarrow12.59)}$ \\
Gemini 3 Flash     & $71.26_{(\uparrow3.97)}$ & $73.39_{(\uparrow8.28)}$ & $72.57_{(\uparrow8.08)}$ & $57.33_{(\downarrow18.01)}$ & $62.46_{(\downarrow12.78)}$ & $60.14_{(\downarrow14.66)}$ & $57.14_{(\downarrow7.91)}$ & $62.27_{(\downarrow4.22)}$ & $60.34_{(\downarrow5.36)}$ & $61.91_{(\downarrow7.32)}$ & $66.04_{(\downarrow2.91)}$ & $64.35_{(\downarrow3.98)}$ \\
Gemini 3 Pro       & $67.29$ & $65.11$ & $64.49$ & $75.34$ & $75.24$ & $74.80$ & $65.05$ & $66.49$ & $65.70$ & $69.23$ & $68.95$ & $68.33$ \\
\midrule
\rowcolor{gray!15} MedForge-Reasoner & $\textbf{99.24}_{(\uparrow31.95)}$ & $\textbf{95.24}_{(\uparrow30.13)}$ & $\textbf{92.86}_{(\uparrow28.37)}$ & $\textbf{99.24}_{(\uparrow23.90)}$ & $\textbf{93.39}_{(\uparrow18.15)}$ & $\textbf{94.86}_{(\uparrow20.06)}$ & $\textbf{99.21}_{(\uparrow34.16)}$ & $\textbf{99.15}_{(\uparrow32.66)}$ & $\textbf{94.09}_{(\uparrow28.39)}$ & $\textbf{99.23}_{(\uparrow30.00)}$ & $\textbf{95.93}_{(\uparrow26.98)}$ & $\textbf{93.94}_{(\uparrow25.61)}$ \\
\midrule
\multicolumn{1}{c}{} &\multicolumn{12}{c}{\textbf{F1 Score}} \\
\midrule
\rowcolor{blue!15} \multicolumn{13}{l}{\textit{Specialized Detectors}} \\
SIDA-7B           & $76.84_{(\uparrow15.02)}$ & $72.47_{(\uparrow9.83)}$ & $67.73_{(\uparrow5.51)}$ & $86.47_{(\uparrow12.25)}$ & $74.06_{(\uparrow2.53)}$ & $69.85_{(\downarrow1.17)}$ & $81.64_{(\uparrow22.17)}$ & $77.28_{(\uparrow19.66)}$ & $72.94_{(\uparrow16.75)}$ & $81.65_{(\uparrow16.48)}$ & $74.60_{(\uparrow10.67)}$ & $70.17_{(\uparrow7.03)}$ \\
SIDA-13B          & \underline{$88.69_{(\uparrow26.87)}$} & $84.95_{(\uparrow22.31)}$ & $80.58_{(\uparrow18.36)}$ & \underline{$90.86_{(\uparrow16.64)}$} & $86.57_{(\uparrow15.04)}$ & $83.17_{(\uparrow12.15)}$ & \underline{$93.18_{(\uparrow33.71)}$} & \underline{$89.46_{(\uparrow31.84)}$} & \underline{$85.87_{(\uparrow29.68)}$} & \underline{$90.91_{(\uparrow25.74)}$} & \underline{$86.99_{(\uparrow23.06)}$} & $83.21_{(\uparrow20.07)}$ \\
FakeVLM           & $85.73_{(\uparrow23.91)}$ & $81.18_{(\uparrow18.54)}$ & $76.29_{(\uparrow14.07)}$ & $88.27_{(\uparrow14.05)}$ & $82.86_{(\uparrow11.33)}$ & $79.16_{(\uparrow8.14)}$ & $90.58_{(\uparrow31.11)}$ & $85.67_{(\uparrow28.05)}$ & $81.84_{(\uparrow25.65)}$ & $88.19_{(\uparrow23.02)}$ & $83.24_{(\uparrow19.31)}$ & $79.10_{(\uparrow15.96)}$ \\
AIGI-Holmes       & $88.57_{(\uparrow26.75)}$ & \underline{$85.84_{(\uparrow23.20)}$} & \underline{$86.68_{(\uparrow24.46)}$} & $90.46_{(\uparrow16.24)}$ & \underline{$88.73_{(\uparrow17.20)}$} & \underline{$84.25_{(\uparrow13.23)}$} & $89.82_{(\uparrow30.35)}$ & $85.97_{(\uparrow28.35)}$ & $83.58_{(\uparrow27.39)}$ & $89.62_{(\uparrow24.45)}$ & $86.85_{(\uparrow22.92)}$ & \underline{$84.84_{(\uparrow21.70)}$} \\
\midrule
\rowcolor{blue!15} \multicolumn{13}{l}{\textit{Generic MLLMs}} \\
Qwen3-VL-Flash     & $32.48_{(\downarrow29.34)}$ & $38.62_{(\downarrow24.02)}$ & $40.26_{(\downarrow21.96)}$ & $55.08_{(\downarrow19.14)}$ & $53.70_{(\downarrow17.83)}$ & $52.69_{(\downarrow18.33)}$ & $63.13_{(\uparrow3.66)}$ & $59.51_{(\uparrow1.89)}$ & $58.78_{(\uparrow2.59)}$ & $50.23_{(\downarrow14.94)}$ & $50.61_{(\downarrow13.32)}$ & $50.58_{(\downarrow12.56)}$ \\
Qwen3-VL-Plus      & $47.73_{(\downarrow14.09)}$ & $46.36_{(\downarrow16.28)}$ & $46.18_{(\downarrow16.04)}$ & $54.70_{(\downarrow19.52)}$ & $53.91_{(\downarrow17.62)}$ & $54.99_{(\downarrow16.03)}$ & $52.42_{(\downarrow7.05)}$ & $54.26_{(\downarrow3.36)}$ & $55.32_{(\downarrow0.87)}$ & $51.62_{(\downarrow13.55)}$ & $51.51_{(\downarrow12.42)}$ & $52.16_{(\downarrow10.98)}$ \\
Gemini 3 Flash     & $24.42_{(\downarrow37.40)}$ & $41.14_{(\downarrow21.50)}$ & $34.78_{(\downarrow27.44)}$ & $70.22_{(\downarrow4.00)}$ & $72.05_{(\uparrow0.52)}$ & $71.11_{(\uparrow0.09)}$ & $70.07_{(\uparrow10.60)}$ & $71.91_{(\uparrow14.29)}$ & $71.34_{(\uparrow15.15)}$ & $54.90_{(\downarrow10.27)}$ & $61.70_{(\downarrow2.23)}$ & $59.08_{(\downarrow4.06)}$ \\
Gemini 3 Pro       & $61.82$ & $62.64$ & $62.22$ & $74.22$ & $71.53$ & $71.02$ & $59.47$ & $57.62$ & $56.19$ & $65.17$ & $63.93$ & $63.14$ \\
\midrule
\rowcolor{gray!15} MedForge-Reasoner & $\textbf{98.86}_{(\uparrow37.04)}$ & $\textbf{93.29}_{(\uparrow30.65)}$ & $\textbf{92.07}_{(\uparrow29.85)}$ & $\textbf{98.86}_{(\uparrow24.64)}$ & $\textbf{92.97}_{(\uparrow21.44)}$ & $\textbf{94.63}_{(\uparrow23.61)}$ & $\textbf{99.21}_{(\uparrow39.74)}$ & $\textbf{99.14}_{(\uparrow41.52)}$ & $\textbf{93.76}_{(\uparrow37.57)}$ & $\textbf{98.98}_{(\uparrow33.81)}$ & $\textbf{95.13}_{(\uparrow31.20)}$ & $\textbf{93.49}_{(\uparrow30.35)}$ \\
\bottomrule
\end{tabular}
}
\caption{\textbf{Main Experiment} - Forgery detection on MedForge-90K dataset. Methods are benchmarked against the Gemini 3 Pro, with arrows indicating performance differences (↑/↓) relative to it. \textbf{Bold} indicates the best result, and \underline{underline} denotes the second-best.}
\label{table: detection_results}
\end{table*}

\subsection{Main Results}
We report the main detection performance in Table~\ref{table: detection_results} with additional results in Appendix~\ref{appendix: additional_results} . To assess performance under both in-domain and out-of-distribution (OOD) conditions, we consider three evaluation settings in Table~\ref{table: detection_results} as follows:

\noindent \textbf{(a) In-Domain:} The detector is trained and tested on the full dataset, covering all forgery types and generator models.

\noindent \textbf{(b) Cross-Model:} To test robustness to unseen generators, we exclude four advanced models from training, Nano-Banana, GPT-Image, Stable Diffusion 3.5 Medium, and Stable Diffusion XL-Inpainting, while evaluating on the default test set.

\noindent \textbf{(c) Cross-Forgery:}  To evaluate generalization to unseen manipulations, the training set excludes lesion implant samples (for both OOD cases, the test data follows the default setting).

Targeting untrainable SOTA commercial MLLM baselines, we simulate the above settings by In Context Learning (ICL) \cite{in-context-leanring}. We design ICL prompts to provide MLLMs with different levels of forgery detection knowledge. Three levels of ICL prompts are customized to match the \textit{In-Domain}, \textit{Cross-Forgery} and \textit{Cross-Model} setting. The \textit{In-Domain} ICL provides detection clues for all manipulation types. The \textit{Cross-Forgery} ICL covers only Lesion Removal Forgeries, and the \textit{Cross-Model} ICL excludes unseen forgery models. See details in Appendix \ref{appendix: generic_llm}.

\begin{table}[t]
\centering
\large
\resizebox{\linewidth}{!}{%
\setlength{\tabcolsep}{4pt} 
\renewcommand{\arraystretch}{1.3}
\begin{tabular}{lccccccccc}
\toprule
\multirow{2}{*}{\textbf{MLLM-as-Judge}} & \multicolumn{4}{c}{\textbf{Gemini 3 Pro (\%)}} & & \multicolumn{4}{c}{\textbf{Qwen3-VL-Plus (\%)}} \\
\cmidrule(lr){2-5} \cmidrule(lr){7-10}
 & LC & VH & MP & \textbf{Avg.} & & LC & VH & MP & \textbf{Avg.} \\
\midrule
SIDA-7B           & 34.6 & 27.9 & 54.3 & 38.9 & & 71.3 & 53.4 & 78.5 & 67.7 \\
SIDA-13B          & 40.2 & 32.6 & 61.6 & 44.8 & & 75.6 & 57.7 & 82.1 & 71.8 \\
FakeVLM           & 48.0 & 53.0 & 45.0 & 48.7 & & 50.0 & 55.0 & 48.0 & 51.0 \\
AIGI-Holmes       & 55.0 & 60.0 & 58.0 & 57.7 & & 58.0 & 62.0 & 57.0 & 59.0 \\
Qwen3-VL-Flash    & 68.2 & 57.3 & 85.9 & 70.5 & & 87.6 & 73.9 & 91.0 & 84.2 \\
Qwen3-VL-Plus     & 69.7 & 59.1 & 84.1 & 71.0 & & 91.9 & \textbf{81.1} & 95.9 & 89.6 \\
Gemini 3 Flash    & 71.2 & 59.0 & 85.0 & 71.7 & & 87.4 & 73.7 & 91.3 & 84.1 \\
Gemini 3 Pro      & \textbf{75.7} & 66.4 & \textbf{86.6} & \textbf{76.2} & & 91.9 & \textbf{81.1} & 95.9 & 89.6 \\
\midrule
\rowcolor{gray!15} Proposed (w/o GSPO) & 71.6 & 66.2 & 83.1 & 73.6 & & 91.1 & 77.7 & 94.2 & 87.7 \\
\rowcolor{gray!15} Proposed (w/ GSPO)  & 71.1 & \textbf{67.4} & 83.2 & 73.9 & & \textbf{93.3} & 79.9 & \textbf{97.5} & \textbf{90.2} \\
\bottomrule
\end{tabular}
}
\caption{\textbf{Evaluation of reasoning quality} via MLLM-as-Judge. We report \textit{Logical Correctness} (LC), \textit{Visual Hallucination} (VH), \textit{Medical Professionalism} (MP), and their \textbf{Average} score in percentage (\%). Gray rows highlight the contribution of Forgery-aware GSPO.}
\label{table: explanation_eval}
\end{table}

As illustrated in Table \ref{table: detection_results}, MedForge-Reasoner achieves SOTA performance across all settings. For the In-Domain setting, our method achieves near-perfect detection, outperforming the strongest specialized detector (SIDA-13B) by over 7.65\% in average accuracy. Notably, MedForge-Reasoner demonstrates significant robustness in OOD scenarios. While specialized detectors and generic MLLMs suffer noticeable performance degradation when facing unseen forgeries or models, our method maintains a substantial lead, surpassing the strongest baselines by 8.2\% in Cross-Forgery and 10.0\% in Cross-Model settings. This suggests that by explicitly training the model to ground its reasoning in visual anomalies (via GSPO), MedForge-Reasoner learns generic traces of tampering (e.g., edge inconsistencies, noise artifacts) rather than overfitting to specific lesion patterns or generator fingerprints. This generalizability suggests that MedForge-Reasoner is applicable in real-world forgery defence.

\subsection{Reasoning Quality}
\label{sec:reasoning_quality}
MedForge-Reasoner provides visually grounded reasoning for detection judgements. Figure \ref{fig:explain-compare} shows a comparison of reasoning outcomes of MedForge-Reasoner and SOTA baselines where MedForge-Reasoner achieves a clear advantage in providing hallucination-free and professional forgery explanations. We further quantitatively evaluate the quality of forgery explanations of all baselines in Table \ref{table: explanation_eval}. For fair comparison, we randomly select 100 forgery samples where all models provide correct detections in the \textit{In-Domain} setting.

The evaluation reveals that, in terms of reasoning quality, MedForge-Reasoner outperforms the best forgery detectors (AIGI-Holmes) by 16.2\% and 31.2\% as measured by the Gemini and Qwen judge. By incorporating the proposed GSPO, our model achieves a leading average Judge Score of 90.2\% under Qwen3-VL-Plus and a competitive 73.9\% under Gemini 3 Pro, outperforming the strongest baseline in the former case. Notably, the GSPO module provides a substantial boost to reasoning quality, increasing the average score by up to 2.5 percentage points compared to the version without GSPO. MedForge-Reasoner demonstrates superior performance in Logical Correctness and Medical Professionalism, while achieving a significant reduction in Visual Hallucination, with scores reaching 79.9\% and 67.4\% under the two judges respectively. This confirms that Forgery-aware GSPO effectively enforces visually grounded reasoning, ensuring the textual output is grounded in visual reality and aligns with medical expertise.

\begin{table}[t]
\centering
\resizebox{\linewidth}{!}{%
\setlength{\tabcolsep}{5pt}
\renewcommand{\arraystretch}{1.2}
\begin{tabular}{l c c c c}
\toprule
\textbf{Setting} & \textbf{Acc (\%)} & \textbf{F1 (\%)} & \textbf{IoU} & \textbf{Judge Score} \\
\midrule
\multicolumn{5}{l}{\textit{\textbf{(A) Contribution of Reasoning Components}}} \\
Binary Classification          & \textbf{99.42} & \textbf{99.31} & - & - \\
w/o Reasoning         & 99.31 & 99.10 & 0.30 & - \\
w/o Bbox Grounding                  & 99.31 & 98.97 & - & 53.9 \\
\rowcolor{gray!15} \textbf{Proposed} & 99.23 & 98.98 & \textbf{0.31} & \textbf{90.2} \\
\midrule
\multicolumn{5}{l}{\textit{\textbf{(B) Efficacy of Optimization Strategies}}} \\
SFT Cold-Start Only            & 98.20 & 98.40 & 0.30 & 87.4 \\
GSPO w/o $R_{\text{bbox}}$     & 99.01 & 98.84 & 0.29 & 89.6 \\
GSPO w/o $R_{\text{form}}$     & 99.13 & 98.98 & 0.30 & 90.1 \\
\rowcolor{gray!15} \textbf{Proposed}   & \textbf{99.23} & \textbf{98.98} & \textbf{0.31} & \textbf{90.2} \\
\midrule
\multicolumn{5}{l}{\textit{\textbf{(C) Scalability across MLLM Backbones}}} \\
InternVL3.5-8B                   & 96.92 & 97.66 & 0.32 & 85.8 \\
Qwen2.5-VL-7B                  & 93.17 & 94.69 & \textbf{0.33} & 80.4 \\
MimoVL-7B                      & 92.26 & 93.91 & 0.32 & 79.1 \\
\rowcolor{gray!15} \textbf{Qwen3-VL-8B (Ours)} & \textbf{99.23} & \textbf{98.98} & 0.31 & \textbf{90.2} \\
\bottomrule
\end{tabular}
}
\caption{Ablation studies on model components, optimization strategies, and backbone architectures.}
\label{table:ablation}
\end{table}

\subsection{Ablation Studies}
In this section, we conduct extensive ablation studies to validate the effectiveness of the proposed architecture and training strategies. To quantify the precision of forgery localization, we additionally report the Intersection over Union (IoU) between the predicted and ground-truth bounding boxes. The ablation studies consist of three parts: \textbf{Part (A)} decomposes the model's response components to assess the necessity of localization and textual rationale; \textbf{Part (B)} isolates the benefits of the specific GSPO training objectives; and \textbf{Part (C)} tests the scalability and robustness of our method across different model architectures. Note that the reasoning quality is evaluated using the Qwen3-VL-Plus judge as described in Section \ref{sec:reasoning_quality}.

\noindent\textbf{Impact of Response Components.} 
As shown in Table~\ref{table:ablation}, the non-interpretable \textit{Binary Classification} and \textit{w/o Reasoning} have the highest detection performance, suggesting that the forgery bounding boxes and textual rationales might slightly interfere with the model's pure decision performance. However, as discussed, black box classification is insufficient for clinical reliability and trustworthy judgment.
Crucially, while the \textit{w/o Bbox Grounding} setting achieves marginally higher accuracy (+0.08\%) than the proposed method, its Judge Score collapses to 53.9\%. This discrepancy reveals that without explicit spatial supervision, the model tends to "hallucinate" justifications, correctly classifying images but for incorrect or non-verifiable reasons. The \textit{Proposed} method achieves the highest Judge Score and IoU with a small accuracy trade-off ($<$0.2\%), showing that MedForge-Reasoner successfully formulates a black-box detection task into interpretable reasoning grounded with factual visual evidence.

\noindent\textbf{Efficacy of GSPO Optimization.} 
Part B disentangles the contributions of our training objectives. 
Although the \textit{SFT Cold-Start} establishes a strong baseline with 98.20\% accuracy, it lags in reasoning quality. Incorporating the proposed GSPO significantly boosts performance. Specifically, removing the spatial reward (\textit{GSPO w/o $R_{\text{bbox}}$}) results in a 0.02 decrease in IoU, verifying that $R_{\text{bbox}}$ is essential for forgery localization. 
Similarly, removing the format reward (\textit{GSPO w/o $R_{\text{form}}$}) leads to a slight degradation in accuracy (99.13\% vs 99.23\%), affecting the logical coherence of the output. 
The full GSPO framework achieves the best balance, yielding the highest Judge Score of \textit{90.2} and Accuracy of \textit{99.23\%}.

\noindent\textbf{Scalability across MLLM Backbones.} 
In Part C, we assess the robustness of our method across different architectures. 
While \textit{InternVL3.5-8B} shows competitive performance (96.92\% Acc), our \textit{Qwen3-VL-8B} based model outperforms it by over 2.3\%. 
Interestingly, although \textit{Qwen2.5-VL-7B} achieves the highest raw IoU (0.33), its reasoning capability is significantly weaker, evidenced by a low Judge Score of 80.4\% and Accuracy of 93.17\%. 
Our proposed method, leveraging the Qwen3-VL backbone, successfully bridges this gap, offering the optimal trade-off between geometric precision and semantic reasoning.

\section{Conclusion}
In this work, we presented a framework to safeguard the trustworthiness of medical imaging against the evolving threat of advanced deepfakes. We established \textbf{MedForge-90K}, the first large-scale medical forgery benchmark with high-fidelity lesion manipulations granularly annotated with expert-aligned reasoning. Addressing the limitations of black-box detectors and hallucination-prone MLLMs, we proposed \textbf{MedForge-Reasoner}, a novel detector capable of \textit{pre-hoc} reasoning. By introducing the Forgery-aware GSPO, we successfully aligned the model's textual outputs with factual visual evidence, explicitly enforcing the detector to localize anomalies before reasoning. Extensive experiments demonstrate that our approach not only achieves state-of-the-art detection performance across unseen forgeries and architectures but also provides clinically rigorous, hallucination-free explanations. We hope this work bridges the gap between AI-driven forgery detection and clinical interpretability, offering a trustworthy solution for high-stakes healthcare environments.


\section{Limitations}
We discuss three main limitations of our work.
First, MedForge-90K currently focuses on three common 2D imaging modalities: chest X-ray, brain MRI, and fundus photography. Although our framework is not modality-specific in principle, extending the benchmark to additional modalities (e.g., CT and ultrasound) and their corresponding forgery patterns would improve coverage of real-world clinical settings.
Second, our reasoning and explanations are generated in English, consistent with most prior work. This choice limits the usability of MedForge-Reasoner in non-English clinical environments. A natural direction for future work is to support multilingual explanations, enabling broader deployment across global healthcare contexts.
Third, while MedForge-Reasoner is designed as a trustworthy medical deepfake detector, it could potentially be misused for malicious purposes, such as improving forgery techniques to evade detection. It is therefore necessary to enforce responsible usage for our released models.

\section{Acknowledgement}
This research is supported by A*STAR, CISCO Systems (USA) Pte. Ltd., and the National University of Singapore under its Cisco-NUS Accelerated Digital Economy Corporate Laboratory (Award I21001E0002), the National Medical Research Council (NMRC) Healthy And Meaningful Longevity - Cognition Grant (NICCOG2024-0028) and the Talent Development Award Project of SSHSPH (26‑0065‑A0001).

\bibliography{custom}

@article{DeepFakesinHealthcare,
author = {Alsaheel, Alaa and Alhassoun, Reem and Alrashed, Reema and Almatrafi, Noura and Almallouhi, Noura and Albahli, Saleh},
year = {2023},
month = {08},
pages = {2461-2482},
title = {Deep Fakes in Healthcare: How Deep Learning Can Help to Detect Forgeries},
volume = {76},
journal = {Computers, Materials Continua},
doi = {10.32604/cmc.2023.040257}
}

@article{forgerymedical,
  title={The Optimal Model for Copy-Move Forgery Detection in Medical Images},
  author={Amiri, Ehsan and Mosallanejad, Ahmad and Sheikhahmadi, Amir},
  journal={Journal of Medical Signals Sensors},
  volume={14},
  number={2},
  pages={5},
  year={2024},
  publisher={Medknow}
}

@article{nanobanana,
  title={Gemini 2.5: Pushing the frontier with advanced reasoning, multimodality, long context, and next generation agentic capabilities},
  author={Comanici, Gheorghe and Bieber, Eric and Schaekermann, Mike and Pasupat, Ice and Sachdeva, Noveen and Dhillon, Inderjit and Blistein, Marcel and Ram, Ori and Zhang, Dan and Rosen, Evan and others},
  journal={arXiv preprint arXiv:2507.06261},
  year={2025}
}

@article{qwenimage,
  title={Qwen-image technical report},
  author={Wu, Chenfei and Li, Jiahao and Zhou, Jingren and Lin, Junyang and Gao, Kaiyuan and Yan, Kun and Yin, Sheng-ming and Bai, Shuai and Xu, Xiao and Chen, Yilei and others},
  journal={arXiv preprint arXiv:2508.02324},
  year={2025}
}

@article{gptimage,
  title={Gpt-4o system card},
  author={Hurst, Aaron and Lerer, Adam and Goucher, Adam P and Perelman, Adam and Ramesh, Aditya and Clark, Aidan and Ostrow, AJ and Welihinda, Akila and Hayes, Alan and Radford, Alec and others},
  journal={arXiv preprint arXiv:2410.21276},
  year={2024}
}

@inproceedings{sd3.5,
  title={Scaling rectified flow transformers for high-resolution image synthesis},
  author={Esser, Patrick and Kulal, Sumith and Blattmann, Andreas and Entezari, Rahim and M{\"u}ller, Jonas and Saini, Harry and Levi, Yam and Lorenz, Dominik and Sauer, Axel and Boesel, Frederic and others},
  booktitle={Forty-first international conference on machine learning},
  year={2024}
}

@article{seedream4,
  title={Seedream 4.0: Toward next-generation multimodal image generation},
  author={Seedream, Team and Chen, Yunpeng and Gao, Yu and Gong, Lixue and Guo, Meng and Guo, Qiushan and Guo, Zhiyao and Hou, Xiaoxia and Huang, Weilin and Huang, Yixuan and others},
  journal={arXiv preprint arXiv:2509.20427},
  year={2025}
}

@InProceedings{sdinpaint,
    author    = {Rombach, Robin and Blattmann, Andreas and Lorenz, Dominik and Esser, Patrick and Ommer, Bj\"orn},
    title     = {High-Resolution Image Synthesis With Latent Diffusion Models},
    booktitle = {Proceedings of the IEEE/CVF Conference on Computer Vision and Pattern Recognition (CVPR)},
    month     = {June},
    year      = {2022},
    pages     = {10684-10695}
}

@article{sdxlinpaint,
  title={Sdxl: Improving latent diffusion models for high-resolution image synthesis},
  author={Podell, Dustin and English, Zion and Lacey, Kyle and Blattmann, Andreas and Dockhorn, Tim and M{\"u}ller, Jonas and Penna, Joe and Rombach, Robin},
  journal={arXiv preprint arXiv:2307.01952},
  year={2023}
}

@misc{flux,
      title={FLUX.1 Kontext: Flow Matching for In-Context Image Generation and Editing in Latent Space},
      author={Black Forest Labs and Stephen Batifol and Andreas Blattmann and Frederic Boesel and Saksham Consul and Cyril Diagne and Tim Dockhorn and Jack English and Zion English and Patrick Esser and Sumith Kulal and Kyle Lacey and Yam Levi and Cheng Li and Dominik Lorenz and Jonas Müller and Dustin Podell and Robin Rombach and Harry Saini and Axel Sauer and Luke Smith},
      year={2025},
      eprint={2506.15742},
      archivePrefix={arXiv},
      primaryClass={cs.GR},
      url={https://arxiv.org/abs/2506.15742},
}

@inproceedings{miccai-detect,
  title={Toward Medical Deepfake Detection: A Comprehensive Dataset and Novel Method},
  author={Li, Shuaibo and Xing, Zhaohu and Wang, Hongqiu and Hao, Pengfei and Li, Xingyu and Liu, Zekai and Zhu, Lei},
  booktitle={International Conference on Medical Image Computing and Computer-Assisted Intervention},
  pages={626--637},
  year={2025},
  organization={Springer}
}

@article{meddetect-from-miccai,
  title={MedNet: Medical deepfakes detection using an improved deep learning approach},
  author={Albahli, Saleh and Nawaz, Marriam},
  journal={Multimedia Tools and Applications},
  volume={83},
  number={16},
  pages={48357--48375},
  year={2024},
  publisher={Springer}
}

@article{fakereasoning,
  title={FakeReasoning: Towards Generalizable Forgery Detection and Reasoning},
  author={Gao, Yueying and Chang, Dongliang and Yu, Bingyao and Qin, Haotian and Chen, Lei and Liang, Kongming and Ma, Zhanyu},
  journal={arXiv preprint arXiv:2503.21210},
  year={2025}
}

@article{veritas,
  title={Veritas: Generalizable deepfake detection via pattern-aware reasoning},
  author={Tan, Hao and Lan, Jun and Tan, Zichang and Liu, Ajian and Song, Chuanbiao and Shi, Senyuan and Zhu, Huijia and Wang, Weiqiang and Wan, Jun and Lei, Zhen},
  journal={arXiv preprint arXiv:2508.21048},
  year={2025}
}

@inproceedings{sida,
  title={Sida: Social media image deepfake detection, localization and explanation with large multimodal model},
  author={Huang, Zhenglin and Hu, Jinwei and Li, Xiangtai and He, Yiwei and Zhao, Xingyu and Peng, Bei and Wu, Baoyuan and Huang, Xiaowei and Cheng, Guangliang},
  booktitle={Proceedings of the Computer Vision and Pattern Recognition Conference},
  pages={28831--28841},
  year={2025}
}

@inproceedings{fakeshield,
    title={FakeShield: Explainable Image Forgery Detection and Localization via Multi-modal Large Language Models},
    author={Xu, Zhipei and Zhang, Xuanyu and Li, Runyi and Tang, Zecheng and Huang, Qing and Zhang, Jian},
    booktitle={International Conference on Learning Representations},
    year={2025}
}

@article{holmes,
  title={AIGI-Holmes: Towards Explainable and Generalizable AI-Generated Image Detection via Multimodal Large Language Models},
  author={Zhou, Ziyin and Luo, Yunpeng and Wu, Yuanchen and Sun, Ke and Ji, Jiayi and Yan, Ke and Ding, Shouhong and Sun, Xiaoshuai and Wu, Yunsheng and Ji, Rongrong},
  journal={arXiv preprint arXiv:2507.02664},
  year={2025}
}

@inproceedings{npr,
  title={Rethinking the up-sampling operations in cnn-based generative network for generalizable deepfake detection},
  author={Tan, Chuangchuang and Zhao, Yao and Wei, Shikui and Gu, Guanghua and Liu, Ping and Wei, Yunchao},
  booktitle={Proceedings of the IEEE/CVF Conference on Computer Vision and Pattern Recognition},
  pages={28130--28139},
  year={2024}
}

@article{system-review,
  title={A systematic literature review on the effectiveness of deepfake detection techniques},
  author={Stroebel, Laura and Llewellyn, Mark and Hartley, Tricia and Ip, Tsui Shan and Ahmed, Mohiuddin},
  journal={Journal of Cyber Security Technology},
  volume={7},
  number={2},
  pages={83--113},
  year={2023},
  publisher={Taylor \& Francis}
}

@inproceedings{mllm-visualhallu,
  title={Visual hallucinations of multi-modal large language models},
  author={Huang, Wen and Liu, Hongbin and Guo, Minxin and Gong, Neil},
  booktitle={Findings of the Association for Computational Linguistics: ACL 2024},
  pages={9614--9631},
  year={2024}
}

@inproceedings{maisi,
  title={Maisi: Medical ai for synthetic imaging},
  author={Guo, Pengfei and Zhao, Can and Yang, Dong and Xu, Ziyue and Nath, Vishwesh and Tang, Yucheng and Simon, Benjamin and Belue, Mason and Harmon, Stephanie and Turkbey, Baris and others},
  booktitle={2025 IEEE/CVF Winter Conference on Applications of Computer Vision (WACV)},
  pages={4430--4441},
  year={2025},
  organization={IEEE}
}

@article{dataaugmentation-motmaed,
  title={Data augmentation using Generative Adversarial Networks (GANs) for GAN-based detection of Pneumonia and COVID-19 in chest X-ray images},
  author={Motamed, Saman and Rogalla, Patrik and Khalvati, Farzad},
  journal={Informatics in medicine unlocked},
  volume={27},
  pages={100779},
  year={2021},
  publisher={Elsevier}
}

@article{securing-forgery,
  title={Securing Healthcare Data Integrity: Deepfake Detection Using Autonomous AI Approaches},
  author={Hsu, Chia-Chi and Tsai, Min-Yan and Yu, Chia-Mu},
  journal={IEEE journal of biomedical and health informatics},
  year={2025},
  publisher={IEEE}
}

@inproceedings{odir,
  title={A benchmark of ocular disease intelligent recognition: One shot for multi-disease detection},
  author={Li, Ning and Li, Tao and Hu, Chunyu and Wang, Kai and Kang, Hong},
  booktitle={International symposium on benchmarking, measuring and optimization},
  pages={177--193},
  year={2020},
  organization={Springer}
}

@article{mimic,
  title={MIMIC-III, a freely accessible critical care database},
  author={Johnson, Alistair EW and Pollard, Tom J and Shen, Lu and Lehman, Li-wei H and Feng, Mengling and Ghassemi, Mohammad and Moody, Benjamin and Szolovits, Peter and Anthony Celi, Leo and Mark, Roger G},
  journal={Scientific data},
  volume={3},
  number={1},
  pages={1--9},
  year={2016},
  publisher={Nature Publishing Group}
}

@misc{mri,
	title={Brain Tumor MRI Dataset},
	url={https://www.kaggle.com/dsv/2645886},
	DOI={10.34740/KAGGLE/DSV/2645886},
	publisher={Kaggle},
	author={Msoud Nickparvar},
	year={2021}
}

@article{multieye,
  title={MultiEYE: Dataset and Benchmark for OCT-Enhanced Retinal Disease Recognition from Fundus Images},
  author={Wang, Lehan and Qi, Chongchong and Ou, Chubin and An, Lin and Jin, Mei and Kong, Xiangbin and Li, Xiaomeng},
  journal={IEEE Transactions on Medical Imaging},
  year={2024},
  publisher={IEEE}
}

@article{yalebrain,
  title={An 11,000-Study Open-Access Dataset of Longitudinal Magnetic Resonance Images of Brain Metastases},
  author={Chadha, Saahil and Weiss, David and Janas, Anastasia and Ramakrishnan, Divya and Hager, Thomas and Osenberg, Klara and Willms, Klara and Zhu, Joshua and Chiang, Veronica and Bakas, Spyridon and others},
  journal={arXiv preprint arXiv:2506.14021},
  year={2025}
}

@misc{gemini3,
  author = {{Google DeepMind}},
  title = {{Gemini 3 Pro Model}},
  howpublished = {\\url{https://deepmind.google/models/gemini/pro/}},
  year = {2025},
  note = {{Accessed on 26 December 2025}}
}

@inproceedings{cva,
  title={Change vector analysis: An approach for detecting forest changes with Landsat},
  author={Malila, William A},
  booktitle={LARS symposia},
  pages={385},
  year={1980}
}

@misc{qwen3vl,
      title={Qwen3-VL Technical Report}, 
      author={Shuai Bai and Yuxuan Cai and Ruizhe Chen and others},
      year={2025},
      eprint={2511.21631},
      archivePrefix={arXiv},
      primaryClass={cs.CV},
      url={https://arxiv.org/abs/2511.21631}, 
}

@article{fakevlm,
  title={Spot the fake: Large multimodal model-based synthetic image detection with artifact explanation},
  author={Wen, Siwei and Ye, Junyan and Feng, Peilin and Kang, Hengrui and Wen, Zichen and Chen, Yize and Wu, Jiang and Wu, Wenjun and He, Conghui and Li, Weijia},
  journal={arXiv preprint arXiv:2503.14905},
  year={2025}
}

@article{diffusion_editors_survey,
  title={Diffusion model-based image editing: A survey},
  author={Huang, Yi and Huang, Jiancheng and Liu, Yifan and Yan, Mingfu and Lv, Jiaxi and Liu, Jianzhuang and Xiong, Wei and Zhang, He and Cao, Liangliang and Chen, Shifeng},
  journal={IEEE Transactions on Pattern Analysis and Machine Intelligence},
  year={2025},
  publisher={IEEE}
}

@article{chen2025med,
  title={Med-Banana-50K: A Cross-modality Large-Scale Dataset for Text-guided Medical Image Editing},
  author={Chen, Zhihui and Feng, Mengling},
  journal={arXiv preprint arXiv:2511.00801},
  year={2025}
}

@inproceedings{in-context-leanring,
  title={A survey on in-context learning},
  author={Dong, Qingxiu and Li, Lei and Dai, Damai and Zheng, Ce and Ma, Jingyuan and Li, Rui and Xia, Heming and Xu, Jingjing and Wu, Zhiyong and Chang, Baobao and others},
  booktitle={Proceedings of the 2024 conference on empirical methods in natural language processing},
  pages={1107--1128},
  year={2024}
}

@inproceedings{divscore,
    title = "{D}iv{S}core: Zero-Shot Detection of {LLM}-Generated Text in Specialized Domains",
    author = "Chen, Zhihui  and
      He, Kai  and
      Huang, Yucheng  and
      Zhu, Yunxiao  and
      Feng, Mengling",
    editor = "Christodoulopoulos, Christos  and
      Chakraborty, Tanmoy  and
      Rose, Carolyn  and
      Peng, Violet",
    booktitle = "Proceedings of the 2025 Conference on Empirical Methods in Natural Language Processing",
    month = nov,
    year = "2025",
    address = "Suzhou, China",
    publisher = "Association for Computational Linguistics",
    url = "https://aclanthology.org/2025.emnlp-main.971/",
    doi = "10.18653/v1/2025.emnlp-main.971",
    pages = "19231--19253",
    ISBN = "979-8-89176-332-6"
}

@article{med-diffu,
  title={Privacy distillation: reducing re-identification risk of multimodal diffusion models},
  author={Fernandez, Virginia and Sanchez, Pedro and Pinaya, Walter Hugo Lopez and Jacenk{\'o}w, Grzegorz and Tsaftaris, Sotirios A and Cardoso, Jorge},
  journal={arXiv preprint arXiv:2306.01322},
  year={2023}
}

@article{synthetic_pneumonia,
  title={A critical assessment of generative models for synthetic data augmentation on limited pneumonia x-ray data},
  author={Schaudt, Daniel and Sp{\"a}te, Christian and von Schwerin, Reinhold and Reichert, Manfred and von Schwerin, Marianne and Beer, Meinrad and Kloth, Christopher},
  journal={Bioengineering},
  volume={10},
  number={12},
  pages={1421},
  year={2023},
  publisher={MDPI}
}

@article{synth-biomarker,
  title={Is synthetic data generation effective in maintaining clinical biomarkers? Investigating diffusion models across diverse imaging modalities},
  author={Hosseini, Abdullah and Serag, Ahmed},
  journal={Frontiers in Artificial Intelligence},
  volume={7},
  pages={1454441},
  year={2025},
  publisher={Frontiers Media SA}
}

@article{Zhang_2026_1, title={MAPS: Multi-Agent Personality Shaping for Collaborative Reasoning}, volume={40}, url={https://ojs.aaai.org/index.php/AAAI/article/view/38669}, DOI={10.1609/aaai.v40i19.38669}, number={19}, journal={Proceedings of the AAAI Conference on Artificial Intelligence}, author={Zhang, Jian and Wang, Zhiyuan and Wang, Zhangqi and Xu, Fangzhi and Lin, Qika and Zhang, Lingling and Mao, Rui and Cambria, Erik and Liu, Jun}, year={2026}, month={Mar.}, pages={16316-16324} }

@article{Zhang_2026_2, title={MARS: Multi-Agent Adaptive Reasoning with Socratic Guidance for Automated Prompt Optimization}, volume={40}, url={https://ojs.aaai.org/index.php/AAAI/article/view/38668}, DOI={10.1609/aaai.v40i19.38668}, number={19}, journal={Proceedings of the AAAI Conference on Artificial Intelligence}, author={Zhang, Jian and Wang, Zhangqi and Zhu, Haiping and Cheng, Kangda and He, Kai and Li, Bo and Lin, Qika and Liu, Jun and Cambria, Erik}, year={2026}, month={Mar.}, pages={16307-16315} }

@article{lin2025has,
  title={Has multimodal learning delivered universal intelligence in healthcare? A comprehensive survey},
  author={Lin, Qika and Zhu, Yifan and Mei, Xin and others},
  journal={Information Fusion},
  volume={116},
  pages={102795},
  year={2025},
  publisher={Elsevier}
}

@article{he2025survey,
  title={A survey of large language models for healthcare: from data, technology, and applications to accountability and ethics},
  author={He, Kai and Mao, Rui and Lin, Qika and others},
  journal={Information Fusion},
  volume={118},
  pages={102963},
  year={2025},
  publisher={Elsevier}
}

@article{li2021knowledge,
  title={Knowledge enhanced lstm for coreference resolution on biomedical texts},
  author={Li, Yufei and Ma, Xiaoyong and Zhou, Xiangyu and Cheng, Pengzhen and He, Kai and Li, Chen},
  journal={Bioinformatics},
  volume={37},
  number={17},
  pages={2699--2705},
  year={2021},
  publisher={Oxford University Press}
}

@article{he2022jcbie,
  title={JCBIE: A joint continual learning neural network for biomedical information extraction},
  author={He, Kai and Mao, Rui and Gong, Tieliang and Cambria, Erik and Li, Chen},
  journal={BMC bioinformatics},
  volume={23},
  number={1},
  pages={549},
  year={2022},
  publisher={Springer}
}

@article{zhang2025patches,
  title={From patches to WSIs: A systematic review of deep Multiple Instance Learning in computational pathology},
  author={Zhang, Yuchen and Gao, Zeyu and He, Kai and Li, Chen and Mao, Rui},
  journal={Information Fusion},
  volume={119},
  pages={103027},
  year={2025},
  publisher={Elsevier}
}

@inproceedings{lin2025self,
  title={Self-supervised quantized representation for seamlessly integrating knowledge graphs with large language models},
  author={Lin, Qika and Zhao, Tianzhe and He, Kai and Peng, Zhen and Xu, Fangzhi and Huang, Ling and Ma, Jingying and Feng, Mengling},
  booktitle={Proceedings of the 63rd Annual Meeting of the Association for Computational Linguistics (Volume 1: Long Papers)},
  pages={13587--13602},
  year={2025}
}

\newpage
\appendix

\section{Expert Acknowledgements}
We would like to express our sincere gratitude to a panel of medical experts for their generous support, valuable clinical insights, and important contributions to the formulation of the Medical Expert Deepfake Guideline. Their expertise greatly strengthened both the medical rigor of this work and the development of the MedForge-90K benchmark. In particular, we would like to acknowledge the following experts:

\begin{itemize}
    \item \textbf{Dr. Yuling Xu} (Ophthalmologist, Guangdong Provincial People's Hospital) offered invaluable clinical insights into fundus image authentication, particularly regarding vascular network logic, the precise morphological and spatial distribution of specific lesions (e.g., macular drusen in AMD and diabetic microaneurysms), and the identification of texture smudging or repetitive artifacts during lesion removal. Dr. Xu's guidance was also instrumental in the construction and validation of the MedForge-90K dataset.
    
    \item \textbf{Dr. Haonan Cai} (Resident Physician, Department of Neurology, Sun Yat-sen Memorial Hospital, Sun Yat-sen University) contributed detailed diagnostic criteria for brain MRI forgeries, with particular emphasis on multi-sequence signal characteristics (T1, T2, and DWI) and anatomical mass effects such as midline shift.
    
    \item \textbf{Dr. Yongzhen Huang} (Resident Physician, Department of General Surgery, Nanfang Hospital of Southern Medical University) provided valuable perspectives on evaluating 3D projection logic in chest X-rays and the biological interconnectivity between lesions and surrounding tissues.
    
    \item \textbf{Dr. Minjing Zhuang} (Ophthalmologist, Guangdong Provincial People's Hospital) contributed important insights into pathological logic in fundus imaging, especially regarding the consistency between primary lesion characteristics and secondary imaging manifestations.
    
    \item \textbf{Dr. Yuhe Lan} (Resident Physician) highlighted the importance of chronological disease progression and emphasized that isolated lesions should remain consistent with secondary clinical and imaging findings.
\end{itemize}

We also extend our heartfelt thanks to the anonymous medical experts whose thoughtful feedback further refined our understanding of medical forgeries and helped shape the foundation of this interpretable benchmark.

\section{Additional Results}
\label{appendix: additional_results}

\subsection{Human Evaluation on Reasoning Quality}
To mitigate potential bias in MLLM-based automatic judges (e.g., stylistic or model-specific preferences), we additionally conducted a single-blind human evaluation. Two annotators with training in medical image annotation single-blindly evaluated the 100 randomly sampled cases using the same rating protocol as the MLLM-as-Judge evaluation in Section~\ref{sec:reasoning_quality}.

As shown in Table \ref{table: human_eval}, our MedForge-Reasoner ranks 3rd best overall, particularly surpassing all specialized deepfake detectors by a large margin. Notably, MedForge-Reasoner achieves the highest score in avoiding Visual Hallucination (76.2\%). While the benchmarked commercial MLLMs (such as the Qwen3-VL series) provide high explanation quality on correctly detected cases, they suffer from poor overall detection accuracy (as previously shown in Table \ref{table: detection_results}). MedForge-Reasoner performs as a balanced solution combining robust detection performance with highly interpretable, hallucination-free reasoning.

\begin{table}[ht]
\centering
\large
\resizebox{0.9\linewidth}{!}{%
\setlength{\tabcolsep}{4pt}
\renewcommand{\arraystretch}{1.3}
\begin{tabular}{lcccc}
\toprule
\multirow{2}{*}{\textbf{Method}} & \multicolumn{4}{c}{\textbf{Human Evaluation (\%)}} \\
\cmidrule(lr){2-5}
 & LC & VH & MP & \textbf{Avg.} \\
\midrule
SIDA-7B           & 52.9 & 52.3 & 53.4 & 52.9 \\
FakeVLM           & 55.2 & 57.2 & 55.3 & 55.9 \\
AIGI-Holmes       & 47.8 & 48.1 & 48.7 & 48.2 \\
Qwen3-VL-Flash    & 75.7 & 75.8 & 75.6 & 75.7 \\
Qwen3-VL-Plus     & \textbf{77.6} & 75.6 & \textbf{75.8} & \textbf{76.3} \\
Gemini 3 Flash    & 64.4 & 63.8 & 64.4 & 64.2 \\
Gemini 3 Pro      & 72.1 & 66.8 & 70.9 & 69.9 \\
\midrule
\rowcolor{gray!15} Proposed (w/o GSPO) & 70.2 & 70.5 & 71.4 & 70.7 \\
\rowcolor{gray!15} Proposed (w/ GSPO)  & 70.3 & \textbf{76.2} & 71.5 & 72.7 \\
\bottomrule
\end{tabular}
}
\caption{\textbf{Human Evaluation on Reasoning Quality.} We report \textit{Logical Correctness} (LC), \textit{Visual Hallucination} (VH), \textit{Medical Professionalism} (MP), and their \textbf{Average} score in percentage (\%). Evaluations were conducted by trained human annotators in a single-blind setting following the same protocol as the MLLM judges.}
\label{table: human_eval}
\end{table}



\subsection{OOD Generalization}

To further evaluate whether detectors generalize beyond the training distribution, rather than relying on shortcut learning or generator-specific artifacts, we further benchmarked MedForge-Reasoner and major baselines on four unseen medical deepfake datasets. These datasets contain manipulations produced by unseen deepfake generators and exhibit artifact patterns that differ from MedForge-90K.

Specifically, we evaluate performance on Med-Diffu~\cite{med-diffu}, Synth-Pneumonia~\cite{synthetic_pneumonia}, Synth-Biomarker~\cite{synth-biomarker}, and Med-Banana~\cite{chen2025med}. Among them, Synth-Pneumonia and Synth-Biomarker are fake-only datasets, we randomly sample real images from MIMIC-CXR to construct a balanced test set with a 1:1 real-to-fake ratio.

As shown in Table~\ref{table:ood_detection}, MedForge-Reasoner achieves the best overall detection performance, maintaining both Accuracy and F1 score above 89\% across all four OOD datasets. In contrast, existing baselines exhibit substantial performance degradation under unseen manipulations. Particularly, FakeVLM collapses to an all-real prediction regime, resulting in a near-zero F1 score.

These results suggest that MedForge-Reasoner captures more generalizable forensic cues of medical image tampering, instead of overfitting to dataset or generator-specific characteristics.

\begin{table}[ht]
\centering
\resizebox{\linewidth}{!}{%
\setlength{\tabcolsep}{4pt}
\renewcommand{\arraystretch}{1.2}
\begin{tabular}{l c c c c}
\toprule
\textbf{Method} & \textbf{Med-Diffu} & \textbf{Synth-Pneu} & \textbf{Synth-Bio} & \textbf{Med-Bana} \\
\midrule
\multicolumn{5}{c}{\textbf{Accuracy (\%)}} \\
\midrule
FakeVLM           & 50.0 & 50.0 & 50.0 & 54.1 \\
AIGI-Holmes       & 51.6 & 51.2 & 49.8 & 49.6 \\
Qwen3-VL-Plus     & 52.5 & 50.5 & 50.2 & 51.0 \\
Gemini 3 Pro      & \underline{90.5} & \underline{80.7} & \underline{74.9} & \underline{70.8} \\
\rowcolor{gray!15}  Proposed  & \textbf{90.7} & \textbf{94.6} & \textbf{92.3} & \textbf{99.6} \\
\midrule
\multicolumn{5}{c}{\textbf{F1 Score (\%)}} \\
\midrule
FakeVLM           & 0.0 & 0.0 & 0.0 & 15.0 \\
AIGI-Holmes       & 63.3 & 63.3 & 62.3 & 61.2 \\
Qwen3-VL-Plus     & 7.7 & 1.0 & 0.4 & 4.2 \\
Gemini 3 Pro      & \textbf{90.1} & \underline{77.5} & \underline{69.3} & \underline{63.6} \\
\rowcolor{gray!15}  Proposed  & \underline{89.8} & \textbf{94.3} & \textbf{91.7} & \textbf{99.6} \\
\bottomrule
\end{tabular}
}
\caption{\textbf{OOD Evaluation on Detection Performance.} We assess generalizability across four unseen medical deepfake datasets.  \textbf{Bold} indicates the best result, and \underline{underline} denotes the second-best.}
\label{table:ood_detection}
\end{table}

\subsection{Failure Mode Analysis}
\label{appendix: failure_mode}

To better understand the remaining errors of MedForge-Reasoner, we conduct a detailed failure mode analysis on the 30,000-sample test set, focusing on three aspects: pathology difficulty, lesion implantation versus removal, and lesion size. Overall, the model exhibits highly consistent performance across edit types and modalities. The accuracy gap between implant and removal forgeries is below 0.2\% in all modalities, suggesting that the detector does not exhibit a strong bias toward either insertion or deletion manipulations. In terms of lesion scale, large lesions ($\geq$8\% of image area) are slightly easier to detect than small/medium lesions ($\leq$8\%), achieving 99.77\% versus 99.49\% accuracy. Across the whole test set, we observe 140 deepfakes misclassified as real (71 implant, 69 removal) and 89 real images misclassified as deepfake, corresponding to an overall error rate of 0.76\%.

Table~\ref{tab:failure_modality_edit} breaks down fake-image accuracy by modality and edit type. The results confirm that the implant--removal gap is consistently small, while the most challenging fake pathology varies across modalities. In particular, MRI remains the most difficult modality, with \textit{Glioma} being the hardest fake category and also having the smallest average GT forgery area ratio.

\begin{table}[ht]
\centering
\small
\setlength{\tabcolsep}{4pt}
\renewcommand{\arraystretch}{1.15}
\begin{tabular}{lccc}
\toprule
\textbf{Modality} & \textbf{Implant} & \textbf{Removal} & \textbf{Area} \\
\midrule
CXR    & 99.37 & 99.47 & 6.05 \\
MRI    & 98.65 & 98.63 & 4.29 \\
Fundus & 99.82 & 99.78 & 5.67 \\
\bottomrule
\end{tabular}
\caption{\textbf{Failure-mode breakdown by modality and edit type.} Values denote fake-image accuracy (\%) for implant/removal and average ground truth forged-area ratio (\%).}
\label{tab:failure_modality_edit}
\end{table}

We further examine the hardest authentic categories that cause false alarms. As shown in Table~\ref{tab:hardest_real}, the dominant source of false positives comes from \textit{Fundus Normal}, which alone contributes 55 errors. This suggests that the detector can occasionally overreact to naturally occurring retinal textures or subtle acquisition variations in healthy fundus images. MRI \textit{Meningioma} and \textit{Pituitary Adenoma} are also among the more difficult authentic categories.

\begin{table}[ht]
\centering
\resizebox{\linewidth}{!}{%
\setlength{\tabcolsep}{4pt}
\renewcommand{\arraystretch}{1.15}
\begin{tabular}{clp{3.1cm}cc}
\toprule
\textbf{Rank} & \textbf{Modality} & \textbf{Pathology} & \textbf{Accuracy (\%)} & \textbf{Errors} \\
\midrule
\rowcolor{gray!15} 1  & Fundus & Normal                      & 97.60 & 55 \\
2  & Fundus & Myopia                      & 98.68 & 1  \\
3  & MRI    & Meningioma                  & 98.73 & 7  \\
4  & MRI    & Pituitary Adenoma           & 98.83 & 7  \\
5  & CXR    & Pneumothorax                & 99.07 & 3  \\
6  & CXR    & Pleural Effusion            & 99.10 & 3  \\
7  & CXR    & Lung Lesion                 & 99.32 & 2  \\
8  & CXR    & Fracture                    & 99.34 & 2  \\
9  & CXR    & Enlarged Cardiomediastinum  & 99.34 & 2  \\
10 & Fundus & Glaucoma                    & 99.40 & 1  \\
\bottomrule
\end{tabular}
}
\caption{\textbf{Top-10 hardest authentic categories} that are misclassified as deepfake (false positive).}
\label{tab:hardest_real}
\end{table}

Table~\ref{tab:hardest_fake} reports the hardest fake categories that are missed by the detector. These errors are concentrated in subtle or anatomically ambiguous pathologies, especially MRI \textit{Glioma}, \textit{Meningioma}, and \textit{Pituitary Adenoma}. We conjecture that such cases are more difficult because their forgeries may preserve coarse anatomical structure while only weakly violating secondary pathological effects, making them challenging even under the proposed localize-then-analyze protocol.

\begin{table}[ht]
\centering
\resizebox{\linewidth}{!}{%
\setlength{\tabcolsep}{4pt}
\renewcommand{\arraystretch}{1.15}
\begin{tabular}{cllcc}
\toprule
\textbf{Rank} & \textbf{Modality} & \textbf{Pathology} & \textbf{Accuracy (\%)} & \textbf{Errors} \\
\midrule
\rowcolor{gray!15} 1  & MRI    & Glioma             & 98.34 & 34 \\
2  & CXR    & Fracture           & 98.39 & 10 \\
3  & MRI    & Meningioma         & 98.68 & 29 \\
4  & CXR    & Pneumothorax       & 98.87 & 7  \\
5  & MRI    & Pituitary Adenoma  & 98.89 & 23 \\
6  & CXR    & Atelectasis        & 99.10 & 6  \\
7  & CXR    & Pneumonia          & 99.38 & 4  \\
8  & CXR    & Consolidation      & 99.39 & 4  \\
9  & Fundus & Glaucoma           & 99.54 & 6  \\
10 & CXR    & Lung Opacity       & 99.54 & 3  \\
\bottomrule
\end{tabular}
}
\caption{\textbf{Top-10 hardest fake categories} that are misclassified as real (false negative).}
\label{tab:hardest_fake}
\end{table}

Overall, these results show that the dominant residual errors are not caused by a systematic failure on one manipulation direction, but are instead concentrated on a small number of subtle authentic fundus images and anatomically challenging MRI forgeries. This also explains why the overall performance remains near-ceiling while still leaving room for improvement on hard boundary cases.

\subsection{Localization IoU Analysis and BBox Diagnostics}
\label{appendix:iou_analysis}

Because our GSPO reward uses one-sided forgery coverage rather than strict IoU, a natural concern is whether the model could exploit the objective by predicting overly large bounding boxes. To directly assess this issue, we report the localization IoU statistics on true-positive forgery test samples in Table~\ref{tab:iou_stats}. Importantly, GSPO improves localization quality rather than collapsing to trivial large-box solutions, which suggests that reward hacking does not dominate in our current setting.

\begin{table}[ht]
\centering
\small
\setlength{\tabcolsep}{4pt}
\renewcommand{\arraystretch}{1.15}
\begin{tabular}{cccccc}
\toprule
\textbf{Mean} & \textbf{Std} & \textbf{Median} & \textbf{95th} & \textbf{$>0.25$} & \textbf{$>0.5/0.75$} \\
\midrule
31.55 & 29.22 & 25.06 & 87.33 & 50.0\% & 27.2\% / 11.7\% \\
\bottomrule
\end{tabular}
\caption{\textbf{Localization IoU statistics} on true-positive forgery test samples ($n=20{,}000$). IoU summary values are reported in percentage.}
\label{tab:iou_stats}
\end{table}
The IoU distribution reveals two important observations. First, localization is often coarse rather than pixel-tight: while the mean IoU is 31.55\% and the median is 25.06\%, half of the correctly detected fake samples still achieve IoU above 0.25. Second, a non-trivial subset of cases is localized very accurately, as reflected by the 95th percentile of 87.33\%. This behavior is consistent with the design of our method: the reward primarily encourages the model to attend to the correct forensic evidence region before reasoning, rather than optimizing for object-detection-style boundary precision.

Finally, although the current results do not indicate systematic reward hacking, the one-sided coverage design may still admit oversized-box solutions in principle. A natural direction for future improvement is to combine the current grounding reward with explicit size-aware regularization, such as area penalty or area-normalized overlap, in order to better suppress trivial large-box predictions while preserving the stability advantages of coverage-based optimization.

\section{MedForge-90K Implementation Details}
\label{sec:appendix_medforge}
To construct high-fidelity and anatomically plausible medical forgeries, we implemented a rigorous pipeline involving automated prompt engineering and diverse image generation models. This section details the specific implementations of the prompt generation, the iterative refinement loop, and the generator models used.

\subsection{Data Collection}
To ensure the authenticity of the source material and the clinical relevance of the forgeries, we curated a diverse collection of 31,990 high-resolution medical images from public benchmarks. As detailed below, our collection spans three imaging modalities and covers 19 distinct pathologies, along with healthy images for each modality.

\paragraph{Chest X-Ray (CXR)}
We sourced frontal-view radiographs from the MIMIC-CXR dataset \cite{mimic}. To facilitate precise lesion removal and implantation, we specifically filtered for scans annotated with exactly one positive pathology. The subset includes 11 thoracic conditions: 
\textit{Atelectasis}, \textit{Cardiomegaly}, \textit{Consolidation}, \textit{Pulmonary Edema}, \textit{Enlarged Cardiomediastinum}, \textit{Rib Fracture}, \textit{Lung Lesion}, \textit{Lung Opacity}, \textit{Pleural Effusion}, \textit{Pneumonia}, and \textit{Pneumothorax}. Healthy control images were selected from the \textit{No Finding} category.

\paragraph{Brain MRI}
Magnetic Resonance Imaging data was sourced from the Brain Tumor Classification dataset \cite{mri} and Yale-Brain \cite{yalebrain}. We focused on contrast-enhanced MRI scans, organizing them into 3 specific tumor typologies: 
\textit{Glioma}, \textit{Meningioma}, and \textit{Pituitary Tumor}. A corresponding set of healthy brain scans was collected under the \textit{Healthy Control} (No Tumor) category to serve as the baseline for tumor implantation tasks.

\paragraph{Fundus Photography}
Retinal images were collected from the ODIR-5K (Ocular Disease Intelligent Recognition) dataset \cite{odir} and MultiEYE \cite{multieye}. We categorized the data into 5 major ocular pathologies based on the diagnostic labels: 
\textit{Age-related Macular Degeneration (AMD)}, \textit{Diabetic Retinopathy}, \textit{Glaucoma}, \textit{Hypertensive Retinopathy}, and \textit{Pathological Myopia}. The \textit{Normal} category was used for healthy reference images.

\paragraph{Preprocessing}
To ensure compatibility with high-fidelity diffusion models, all raw images underwent a standardization pipeline. Images were resized and padded to a uniform resolution of $1024 \times 1024$ pixels, strictly preserving the original aspect ratio to maintain anatomical integrity before being fed into the forgery generation loop.

\paragraph{Dataset inventory.}
Table~\ref{tab:medforge90k_editor_stats} summarizes verified RGB image counts from the MedForge-90K release manifest. Across authentic and forged splits, the benchmark comprises \textbf{95,276} images in total (31,990 real \& 63,286 deepfake).

\begin{table*}[htbp]
\centering
\small
\setlength{\tabcolsep}{5pt}
\renewcommand{\arraystretch}{1.18}
\begin{tabular}{@{}l l r r r@{}}
\toprule
\rowcolor{gray!12}\textbf{Deepfake Model} & \textbf{Dataset ID} & \textbf{Total} & \textbf{Implant} & \textbf{Remove} \\
\midrule
Stable Diffusion Inpainting (v1.5) & \texttt{stable-diffusion-inpainting} & 6,303 & 3,239 & 3,064 \\
Stable Diffusion XL Inpainting 0.1 & \texttt{stable-diffusion-xl-1.0-inpainting-0.1} & 6,380 & 3,273 & 3,107 \\
FLUX.1-dev & \texttt{flux.1-dev} & 6,399 & 3,273 & 3,126 \\
Stable Diffusion 3.5 Large & \texttt{stable-diffusion-3.5-large} & 6,383 & 3,273 & 3,110 \\
Stable Diffusion 3.5 Medium & \texttt{stable-diffusion-3.5-medium} & 6,383 & 3,273 & 3,110 \\
GPT-Image & \texttt{gpt} & 6,340 & 3,223 & 3,117 \\
Gemini 2.5 Flash Image & \texttt{gemini} & 6,413 & 3,273 & 3,140 \\
Qwen-Image-Edit & \texttt{qwen} & 6,347 & 3,213 & 3,134 \\
Seedream 4.0 & \texttt{seeddream} & 6,155 & 3,131 & 3,024 \\
Step1X-Edit & \texttt{step1x-edit} & 6,183 & 3,223 & 2,960 \\
\midrule
\rowcolor{gray!18}\textbf{All Models (forged)} & --- & \textbf{63,286} & \textbf{32,394} & \textbf{30,892} \\

\rowcolor{gray!18}\textbf{Real (authentic)} & --- & \textbf{31,990} & --- & --- \\
\rowcolor{gray!18}\textbf{Total} & --- & \textbf{95,276} & --- & --- \\
\bottomrule
\end{tabular}
\caption{\textbf{MedForge-90K image inventory.} Per-editor RGB forgeries after the Writer--Editor--Diagnoser loop; \textit{Implant} and \textit{Remove} match the \texttt{*-edit} and \texttt{*-remove} splits. Deepfake models follow the presentation order in Appendix~\ref{sec:appendix_forgery_gen} (inpainting $\rightarrow$ advanced diffusion $\rightarrow$ proprietary APIs), with Step1X-Edit listed last as an additional open instruction editor. Footer rows report forged aggregates, the authentic split total, and the full RGB inventory (\textit{Implant}/\textit{Remove} are undefined for real and for the dataset-wide total).}
\label{tab:medforge90k_editor_stats}
\end{table*}

\subsection{Forgery Prompt Generation}
To guide the image editing models in performing precise lesion manipulation, we utilize a Large Language Model (Gemini 2.5 Pro) acting as the \textit{Writer}. The goal is to translate medical tasks (e.g., ``Implant Pleural Effusion'' or ``Remove Brain Tumor'') into natural language instructions understandable by text-guided image editing models.

The generation process adheres to three critical constraints to ensure the output is realistic and undetectable as a deepfake:
\begin{enumerate}
    \item \textbf{Fidelity Preservation:} The prompt must explicitly instruct the editor to preserve original image noise, grain texture, and contrast, avoiding alterations to image.
    \item \textbf{Negative Rules:} We enforce strict negative constraints, forbidding the addition of text, labels, or unnatural sharp boundaries.
    \item \textbf{Minimal Change Principle (Counterfactual Minimality):} The prompt emphasizes modifying \textit{only} the pixels necessary for the pathology, leaving the background and surrounding anatomy untouched.
\end{enumerate}

For \textbf{Lesion Implant}, the system instruction provided to the \textit{Writer} is:
\begin{tcolorbox}[colback=gray!10, colframe=gray!50, title=Lesion Implant Prompt Generation]
\small
``You are a medical image editing expert. Generate a clear, concise prompt to edit a normal [Modality] image to show [Disease]. [...]
Critical Constraints - Fidelity Preservation: Preserve original image noise, grain texture, and contrast. Do not alter device artifacts...
Critical Constraints - Minimal Change Principle: Only modify areas directly related to [Disease]. Keep all other anatomical structures unchanged...
Add a clear warning: do not edit any element other than adding the disease feature. Keep everything else in the image exactly the same.''
\end{tcolorbox}

For \textbf{Lesion Removal}, the instruction shifts to describing the removal of specific anomalies without leaving inpainting traces:
\begin{tcolorbox}[colback=gray!10, colframe=gray!50, title=Lesion Removal Prompt Generation]
\small
``Generate a clear, concise prompt to edit a [Modality] image showing [Disease] to make it appear normal (healthy). [...]
The prompt should not include any medical-related terms... and only describe the direct modifications in the simplest way (e.g., 'delete the white rounded shape').
The prompt should focus on locating the lesion and describing the boundary... and not specify the replacement content.''
\end{tcolorbox}

\subsection{Forgery Prompt Refinement}
Forgery generator could fail to produce medically accurate or visually seamless results. To address this, we implement a \textit{Writer-Editor-Diagnoser} refinement loop. 

\paragraph{The Verification Loop (Diagnoser)}
In each iteration, the \textit{Editor} generates a candidate image. A \textit{Diagnoser} (Gemini 3 Pro) then performs a pixel-level side-by-side comparison between the original and the forged image to ensure the pathology is added/removed correctly without affecting the background. The specific instruction used for this verification is:

\begin{tcolorbox}[colback=gray!10, colframe=gray!50, title=Forgery Verification Instruction]
\small
``You are a medical image verification expert. You are given two images:
1. Original image ... 2. Edited image ...
Your task is to verify the editing quality by comparing the two images side-by-side.

Critical Verification - Minimal Change Principle:
Compare the original and edited images carefully. The editing should only modify the disease-related regions. Check:
- Are non-disease areas (background, other anatomical structures, imaging artifacts) identical?
- Does the edited version preserve the exact same imaging characteristics (noise, grain, contrast)?

Check these aspects:
1. Has disease: Does the edited image show signs of [Disease]?
2. Structure reasonable: Are the anatomical structures reasonable and correct?
3. Looks realistic: Does the edited image look like a real medical image?
4. Minimal changes preserved: Are changes limited only to disease areas?

Return your evaluation in this JSON format: \{ "qualified": true/false, "reason": "..." \}''
\end{tcolorbox}

\paragraph{The Prompt Refinement (Writer)}
If the verification fails (e.g., due to artifacts or incorrect anatomy), the execution history and failure reasons are fed back to the \textit{Writer}. The \textit{Writer} is then prompted to analyze the previous failures and generate an improved prompt:

\begin{tcolorbox}[colback=gray!10, colframe=gray!50, title=Forgery Prompt Refinement Instruction]
\small
``You are a medical image editing expert. Multiple previous editing attempts have failed. You need to analyze ALL previous attempts and generate a BETTER prompt.

History of all previous attempts: 
[History Log]

Looking at the ORIGINAL image and analyzing the patterns of failures above, generate an IMPROVED editing prompt.
ANALYSIS REQUIREMENTS:
1. Identify common issues across multiple attempts
2. Learn from what didn't work in previous rounds
3. Avoid repeating the same mistakes

[...] (Standard constraints on Fidelity Preservation and Minimal Change Principle are repeated here)

Return ONLY the editing prompt in English, no explanations.''
\end{tcolorbox}

This loop repeats for up to 5 rounds. Only images that pass the strict verification criteria (``qualified'': true) are included in the MedForge-90K.

\subsection{Forgery Generation}\label{sec:appendix_forgery_gen}
Once the prompts are refined and validated, we employ a diverse ensemble of 10 state-of-the-art image editing and generation models to construct the final MedForge-90K dataset. Using a wide range of architectures prevents the detector from overfitting to specific generator artifacts (e.g., specific noise patterns of a single diffusion model). The models utilized are categorized as follows:

\paragraph{Diffusion-based Inpainting Models}
These models require a mask (derived from the Nano-Banana coordinates) and the refined text prompt to regenerate specific regions.
\begin{itemize}
    \setlength{\itemsep}{0pt}
    \setlength{\parskip}{0pt}
    \item \textbf{Stable Diffusion Inpainting (SD-v1.5)}: A baseline latent diffusion model specialized for mask-based editing \cite{sdinpaint}.
    \item \textbf{Stable Diffusion XL (SDXL) Inpainting 0.1}: A larger scale model (2.6B parameters) capable of generating higher resolution details and better texture matching in medical scans \cite{sdxlinpaint}.
\end{itemize}

\paragraph{Advanced Diffusion-based Image Editing Models}
These models perform instruction-based editing without needing explicit masks, relying on the refined prompts to localize and modify content.
\begin{itemize}
    \setlength{\itemsep}{0pt}
    \setlength{\parskip}{0pt}
    \item \textbf{FLUX.1-dev}: A 12B parameter rectified flow transformer model. It is chosen for its superior prompt adherence and ability to generate high-frequency details (noise/grain) crucial for medical realism \cite{flux}.
    \item \textbf{Stable Diffusion 3.5 Large}: The latest Multimodal Diffusion Transformer (MMDiT) from Stability AI, offering state-of-the-art conceptual understanding of complex prompts \cite{sd3.5}.
    \item \textbf{Stable Diffusion 3.5 Medium}: A distilled version of SD3.5, providing a variation in generation artifacts to test detector robustness against model compression traces \cite{sd3.5}.
\end{itemize}

\paragraph{MMDiT Image Editing Models}
We also utilize closed-source or specialized APIs to capture the distribution of commercial deepfake tools.
\begin{itemize}
    \setlength{\itemsep}{0pt}
    \setlength{\parskip}{0pt}
    \item \textbf{GPT-Image}: Accessed via OpenAI API. Known for high semantic understanding, used primarily for complex lesion removal tasks where context reasoning is required \cite{gptimage}.
    \item \textbf{Gemini-2.5-Flash-Image (Nano-Banana)}: Accessed via Google GenAI API. Utilized for its strong instruction-following capabilities in medical contexts \cite{nanobanana}.
    \item \textbf{Qwen-Image-Edit}: Based on the Qwen-Image architecture, this model integrates visual understanding with generation, allowing for precise editing based on visual cues \cite{qwenimage}. It stands out as the SOTA open-source image editor currently, making it crucial to evaluate on medical forgery detection. 
    \item \textbf{Seedream 4.0}: A high-performance multimodal image generation model designed for high-consistency semantic editing, minimizing changes to the background \cite{seedream4}. Seedream 4.0 is pretrained on billions of text-image pairs spanning diverse taxonomies and knowledge-centric concepts, suitable for high-quality medical forgeries.
\end{itemize}

This ensemble ensures that MedForge-90K covers open-source latent diffusion models and proprietary MMDiT-based generators, representing a comprehensive threat landscape.

\subsection{Reasoning Annotation}
\label{sec:reasoning_annotation}

To equip MedForge-90K with granular and clinically grounded explanations, we developed an automated annotation pipeline utilizing advanced MLLMs (Gemini 2.5 Pro). Unlike standard captioning tasks, our pipeline employs a \textbf{Hierarchical Guideline-Driven Reasoning} strategy. This mechanism enforces the model to scrutinize images through a three-tiered cognitive framework derived directly from our expert guidelines (detailed in Section \ref{sec:guidelines}):
\begin{itemize}
    \setlength{\itemsep}{0pt}
    \setlength{\parskip}{0pt}
    \item \textbf{Level 1: Image Physics \& Texture.} Detecting low-level anomalies such as "sticker" artifacts, unnatural noise distribution, or inpainting smudges that violate the physical properties of medical imaging.
    \item \textbf{Level 2: Anatomical Structure.} Verifying morphological correctness, such as the continuity of vascular networks in fundus photography or the symmetry of gyri in brain MRI.
    \item \textbf{Level 3: Pathological Logic.} Checking high-level biological interconnectivity to ensure lesions exhibit necessary secondary effects (e.g., mass effect, edema, chronological progression) rather than in biological isolation.
\end{itemize}

\paragraph{Annotation for Authentic Images}
For real images, the pipeline shifts to validating \textit{Biological Consistency}. The prompt directs the MLLM to confirm the \textit{satisfaction} of the hierarchical logic—verifying that noise patterns are stochastic, anatomy is continuous, and pathological signs follow a natural progression. This ensures the detector learns the logic of authenticity, distinct from the features of forgery.
\paragraph{Annotation for Forged Images}
For images in the \textit{Lesion Implant} and \textit{Removal} categories, the annotation is spatially grounded using the ground-truth manipulation mask for hierarchical analysis:
\begin{enumerate}
    \setlength{\itemsep}{0pt}
    \setlength{\parskip}{0pt}
    \item \textbf{Bbox Extraction:} Bounding box coordinates $\mathbf{b} = [x_1, y_1, x_2, y_2]$ are extracted from the binary manipulation mask.
    \item \textbf{Hierarchical Prompting:} The prompt that explicitly informs the MLLM of the forgery location. Crucially, we inject the specific \textit{Expert Forgery Guidelines} into the prompt context. The model is instructed to analyze the image area within $\mathbf{b}$ specifically for violations across the previously defined logic levels.
\end{enumerate}

\begin{tcolorbox}[colback=gray!10, colframe=gray!50, title=Hierarchical Forgery Reasoning Prompt Template]
\small
\textbf{System:} This is a medical deepfake image. The bounding box $[y_{min}, x_{min}, y_{max}, x_{max}]$ indicates the location of the deepfake region.
\textbf{Task:} Analyze why this image is a deepfake by systematically applying the \textit{Medical Deepfake Detection Guidelines} provided below.
\textbf{Requirements:}
1. \textbf{Location:} Output the coordinates in \texttt{<box>} format.
2. \textbf{Description:} Briefly describe the image modality and features.
3. \textbf{Key Explanation:} Identify anomalies following the hierarchical logic:
   - \textit{Physics:} Are there noise/texture artifacts or sharp boundaries?
   - \textit{Anatomy:} Are structures morphologically incorrect?
   - \textit{Pathology:} Are biological secondary signs missing (e.g., lack of mass effect)?
4. \textbf{Conclusion:} Definitive statement of forgery.
\textbf{Context:} [Injected Modality-Specific Guidelines from Section \ref{sec:guidelines}]
\end{tcolorbox}

\subsection{Medical Deepfake Detection Guidelines}
\label{sec:guidelines}

To ensure the reasoning annotations described above align with clinical expertise, we formulated a comprehensive set of detection criteria. These guidelines serve as the "ground truth logic" injected into the annotation prompts.

\subsubsection{General Principles}
This section applies to all image modalities, focusing on the failure of AI forgery to replicate biological interconnectivity and physical consistency.

\noindent \textbf{Biological Plausibility \& Secondary Effects}
    \begin{itemize}
    \setlength{\itemsep}{0pt}
    \setlength{\parskip}{0pt}
        \item \textbf{Mass Effect Absence:} Real lesions are physical objects that displace tissue. Reject if a space-occupying lesion exists without corresponding compression, displacement, or midline shift.
        \item \textbf{Lack of Host Reaction:} The body reacts to pathology. Reject if an aggressive lesion appears "isolated" with a sharp boundary and no surrounding edema or infiltration.
        \item \textbf{Chronological Inconsistency:} Diseases follow a timeline. Reject if late-stage features appear without precursor signs (e.g., neovascularization without ischemia).
    \end{itemize}
\textbf{Image Physics \& Texture Consistency}
    \begin{itemize}
    \setlength{\itemsep}{0pt}
    \setlength{\parskip}{0pt}
        \item \textbf{The "Sticker" Artifact:} Reject if the lesion-background interface is unnaturally sharp, lacking the gradual transition zone of biological tissues.
        \item \textbf{Noise Distribution Analysis:} Reject if the noise pattern (grain) within the lesion is significantly smoother or different in texture compared to the surrounding unaffected tissue.
        \item \textbf{Inpainting Artifacts:} In removal cases, look for "smudging," blurring, or repetitive cloning patterns that disrupt natural stochastic texture.
    \end{itemize}

\subsubsection{Modality-Specific Principles}
These criteria address the specific anatomical and structural logic required for each imaging type.

\paragraph{Brain MRI}
\begin{itemize}
    \setlength{\itemsep}{0pt}
    \setlength{\parskip}{0pt}
    \item \textbf{Anatomical Logic:} Sulci adjacent to a mass should be effaced. The ventricular system must be symmetrical unless physically displaced. Large unilateral masses must cause a contralateral midline shift.
    \item \textbf{Signal Intensity:} Peritumoral edema must follow correct signal intensity (e.g., Hyperintense on T2/FLAIR). Lesions must match specific signatures (e.g., Meningiomas require a "Dural Tail").
    \item \textbf{Multi-Sequence Consistency:} Lesion appearance must logically translate across sequences (e.g., fluid is bright on T2, dark on T1).
\end{itemize}

\paragraph{Fundus Photography}
\begin{itemize}
    \setlength{\itemsep}{0pt}
    \setlength{\parskip}{0pt}
    \item \textbf{Vascular Logic:} Vessels must taper gradually from the optic disc to the periphery without discontinuities. The Artery/Vein (A/V) ratio must be consistent.
    \item \textbf{Lesion Distribution:} Diabetic lesions usually spare the extreme periphery initially. Drusen must be concentrated in the Macula. Macular exudates should form a "Star" pattern due to Henle’s fiber layer.
    \item \textbf{Global Physics:} The image must exhibit natural vignetting (e.g., posterior pole brighter than periphery).
\end{itemize}

\paragraph{Chest X-Ray (CXR)}
\begin{itemize}
    \setlength{\itemsep}{0pt}
    \setlength{\parskip}{0pt}
    \item \textbf{3D Projection Logic:} Lung markings must correctly overlap with ribs/heart. Skeletal structures (rib count, clavicle shape) must be anatomically correct.
    \item \textbf{Density Gradient:} Adherence to the density ladder (Air $<$ Fat $<$ Bone). Vascular markings should be more prominent in lower zones.
    \item \textbf{Secondary Signs:} Atelectasis must show volume loss (elevated diaphragm). Cardiomegaly should manifest with pulmonary congestion.
\end{itemize}

\section{Experiment Settings}
\label{appendix: settings}
\subsection{Baselines: Generic MLLM}
\label{appendix: generic_llm}
To evaluate the zero-shot and in-context reasoning capabilities of state-of-the-art models in medical deepfake detection, we employ four representative Multi-modal Large Language Models (MLLMs). These include the \textbf{Qwen3-VL} series and the \textbf{Gemini 3} series, known for SOTA image understanding and reasoning abilities.
\subsubsection{Model Settings}
All models are accessed via their respective official APIs to ensure reproducibility. The specific models and their configurations are as follows:
\begin{itemize}
    \setlength{\itemsep}{0pt}
    \setlength{\parskip}{0pt}
\item \textbf{Qwen3-VL-Flash \& Qwen3-VL-Plus}: Accessed via Qwen API.
\item \textbf{Gemini 3 Flash \& Gemini 3 Pro}: Accessed via the Google Generative AI (GenAI) SDK. 
\end{itemize}
For all API calls, we set the \texttt{temperature} to $0.1$ to minimize stochasticity and encourage deterministic, logical outputs. The maximal number of output tokens is set to $1024$ to accommodate the detailed judgement explanations.

\subsubsection{In-Context Learning Prompts}
We design three distinct levels of In-Context Learning (ICL) prompts to evaluate the model's generalization capability across different forensic scenarios. These prompts are generated using a "Forensics Expert" agent (powered by Gemini 3 Pro) based on a selected set of real and manipulated medical examples.
\begin{enumerate}
    \setlength{\itemsep}{0pt}
    \setlength{\parskip}{0pt}
\item \textbf{In-Domain ICL Prompt}: Contains comprehensive guidance covering all available modalities (CXR, MRI, Fundus) and all generator architectures (SD, Flux, GANs). It serves as the upper bound for model performance when full forensic knowledge is available.
\item \textbf{Cross-Model ICL Prompt}: Excludes specific generative models (e.g., Stable Diffusion, GPT-based generators) from the context to test if the MLLM can generalize forensic principles to "unseen" generator artifacts.
\item \textbf{Cross-Forgery ICL Prompt}: Focuses primarily on one type of manipulation (e.g., lesion removal) while excluding others (e.g., implants/edits), evaluating the model's ability to identify fundamental biological inconsistencies regardless of the forgery task.
\end{enumerate}

\begin{figure*}[h]
    \centering
    \includegraphics[width=1\linewidth]{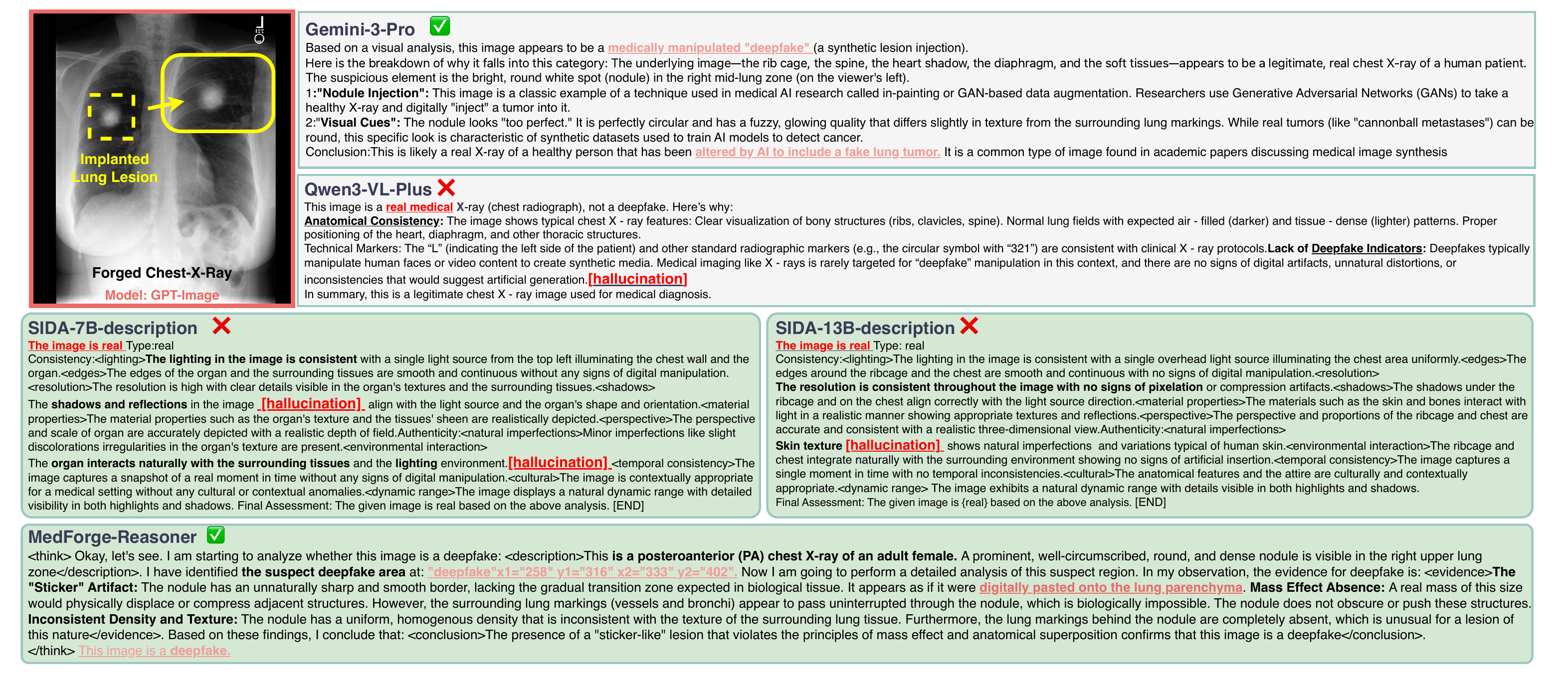}
    \caption{\textbf{Qualitative Forgery Explanation Comparison.} Baselines fail due to severe hallucinations (SIDA citing ``skin texture'') or missed diagnoses. While Gemini-3-Pro correctly detects the forgery using general visual clues, MedForge-Reasoner delivers superior \textit{clinically rigorous rationale}, explicitly grounding the verdict in anatomical logic (e.g., ``absence of mass effect'') rather than generic visual analysis.}
    \label{fig:explain-compare}
\end{figure*}

\subsection{Baselines: Specialized Detectors}
To ensure a fair comparison, all specialized baseline detectors were trained on the MedForge-90K training set. We followed the official implementations and recommended hyper-parameters provided by the respective authors, adapting them to medical forgery detection. All models were trained on 8 NVIDIA H100 80G GPUs, requiring around 10-15 hours per model.

\paragraph{SIDA-7B \& 13B} 
SIDA~\cite{sida} is an MLLM-based detector designed for forgery detection and localization. We utilized the LLaVA-v1.5 [7B/13B] as the backbone.
\begin{itemize}
    \setlength{\itemsep}{0pt}
    \setlength{\parskip}{0pt}
    \item \textbf{Training Stage:} We performed default LoRA training (rank=128 alpha256) on the MedForge-90K SFT split.
    \item \textbf{Hyper-parameters:} Following the same setting as proposed detector, SIDA 7/13B were trained for 10 epochs with a total batch size of 8. We used the AdamW optimizer with a learning rate of 2e-5.
    \item \textbf{Original Setting:} Following the original implementation, the input resolution was set to 336$\times$336, and the prompt followed the "instruction-reasoning-label" format as described in the original paper.
\end{itemize}

\paragraph{FakeVLM}
FakeVLM \cite{fakevlm}, a specialized large multimodal model designed for both general synthetic image and DeepFake detection.
\begin{itemize}
    \setlength{\itemsep}{0pt}
    \setlength{\parskip}{0pt}
    \item \textbf{Backbone:} We employed llava-1.5-7b as the detector backbone, following the original implementation
    \item \textbf{Training Stage:} The model underwent 10 stages of LoRA training (rank=128 alpha256), based on forgery label and textual descriptions formulated from the MedForge-90K SFT set.
\end{itemize}

\paragraph{AIGI-Holmes}
AIGI-Holmes~\cite{holmes} utilizes a multi-stage framework consisting of a CLIP-based forgery Visual Expert and subsequent LLM reasoning explainer.
\begin{itemize}
    \item \textbf{Backbone:} We employed CLIP and NPR network as the Visual Expert, and llava-v1.6-mistral-7b-hf as the LLM backbone following the original setting.
    \item \textbf{Training Stage:} The visual experts are trained following default configuration. The LLM module underwent 10 stages of LoRA training (rank=128 alpha256), based on the MedForge-90K SFT set.
\end{itemize}

\subsection{Evaluation Metrics}
\label{appendix: judge_details}
We report the detection performance using Accuracy and F1 Score metrics. These are standard metrics calculated based on True Positives (TP), True Negatives (TN), False Positives (FP), and False Negatives (FN). The formulas are defined below:
\begin{equation}
\small
    \text{Accuracy} = \frac{TP + TN}{TP + TN + FP + FN}
\end{equation}

\begin{equation}
\small
    \text{F1} = \frac{2 \cdot TP}{2 \cdot TP + FP + FN}
\end{equation}
In the main experiment (Table \ref{table: detection_results}), we report performance broken down by category. To ensure clarity, we define the specific positive and negative classes used for calculating metrics in each column:

\begin{itemize}
    \setlength{\itemsep}{0pt}
    \setlength{\parskip}{0pt}
    \item \textbf{Real:} This measures the model's ability to identify authentic images. Here, the positive class is the \textit{Real} image, and the negative class includes all \textit{Fake} images (comprising both Lesion Implant and Lesion Removal).
    \item \textbf{Forgery Implant:} This measures the model's ability to distinguish implanted lesions from healthy tissue. Here, the positive class is the \textit{Lesion Implant} forgery, and the negative class is the \textit{Real} image. Lesion Removal samples are excluded from this calculation to isolate the performance on implantation.
    \item \textbf{Forgery Remove:} This measures the model's ability to detect erased lesions. Here, the positive class is the \textit{Lesion Removal} forgery, and the negative class is the \textit{Real} image. Lesion Implant samples are excluded.
\end{itemize}

To quantitatively assess the quality of the generated forensic reasoning, we employ a reference-based evaluation protocol using state-of-the-art MLLMs (Qwen3-VL-Plus and Gemini 3 Pro) as impartial judges. Unlike standard n-gram metrics (e.g., BLEU, ROUGE) which fail to capture semantic consistency in medical diagnostics, our MLLM-as-Judge approach evaluates the \textit{factual consistency} between the model's generated rationale and the Ground Truth forgery reasoning.
As defined in our evaluation script, the judge scores each response on a scale of 1 to 10, which is then converted to a 100\% scale for reporting. The MLLM-as-Judge is based on three distinct criteria:

\begin{enumerate}    
    \setlength{\itemsep}{0pt}
    \setlength{\parskip}{0pt}
\item \textbf{Logical Correctness:} Evaluates whether the assistant's reasoning follows a sound forensic process. It rewards responses that arrive at the correct conclusion through valid deduction, rather than lucky guesses.
\item \textbf{Visual Hallucination:} Measures the faithfulness of the description to the visual reality. A high score indicates the model describes only features present in the Ground Truth (e.g., specific bbox locations, noise patterns), while a low score indicates the fabrication of non-existent features.
\item \textbf{Medical Professionalism:} Assesses whether the terminology (e.g., "mass effect," "vascular continuity") and diagnostic logic align with the provided expert medical guidelines.
\end{enumerate}

\paragraph{Judge Prompt}
To ensure objectivity, the judge is provided with the specific role of a "Medical Image Forensics Expert." The exact prompt used in our evaluation pipeline is presented below:

\begin{tcolorbox}[colback=gray!10, colframe=gray!50, title=MLLM-as-Judge System Prompt, breakable]
\small
\textbf{Role:} You are a Medical Image Forensics Expert acting as an impartial judge. Your expertise covers Radiology (MRI, CXR) and Ophthalmology (Fundus), specifically in identifying AI-generated (Deepfake) anomalies versus real pathological features.

\textbf{Task:} Please examine the provided text responses and serve as an unbiased judge in assessing the quality of a forensic analysis from an AI assistant. You will evaluate how well the assistant identifies and explains the forensic nature of the image manipulation based on professional medical imaging standards, compared to a Ground Truth reference.

\textbf{Input Data:}

\textbf{Assistant Response:} The forensic analysis provided by the AI assistant for evaluation.
\textbf{Ground Truth Information:} The definitive expert reference explanation for the manipulations present in the image.
\textbf{Evaluation Focus:}
Your evaluation should focus exclusively on the content and factual correctness of the assistant's response compared to the Ground Truth.

DO NOT reward for tedious and verbose responses.
DO focus on whether the assistant correctly identified the same forensic anomalies, biological evidence, and medical logic as described in the Ground Truth. Reward the response outputting correct bbox coordinate.
\textbf{Evaluation Criteria:}

\textbf{Logical Correctness:} Whether the assistant's reasoning follows a sound forensic process and arrives at the correct conclusion.
\textbf{Visual Hallucination:} Whether the assistant's verbal description of the anomalies matches the ground truth or fabricates nonexistent features/locations.
\textbf{Medical Professionalism:} Whether the terminology and medical logic used in the text align with expert guidelines.
\end{tcolorbox}

\subsection{MedForge-Reasoner Training}
\label{appendix:training_details}

\paragraph{SFT Cold-Start Stage.}
We utilize the Qwen3-VL-8B-Instruct as the backbone model. We employ LoRA (Low-Rank Adaptation) for parameter-efficient fine-tuning, targeting all linear modules with a rank $r=128$ and alpha $\alpha=256$. The model is trained for 10 epochs using the AdamW optimizer with a learning rate of $1 \times 10^{-4}$ and a cosine decay scheduler (warmup ratio set to 0.05). The training uses a global batch size of 512 (per-device batch size 16 with gradient accumulation) and bfloat16 precision. The maximum sequence length is set to 2048 to accommodate detailed reasoning chains.

\paragraph{Forgery-aware GSPO Stage.}
We initialize the model with the SFT checkpoint and align it using the Group Sequence Policy Optimization (GSPO) framework. The model is trained for 1 epoch with a reduced learning rate of $1 \times 10^{-6}$. We set the group size $G=8$ to sample diverse reasoning paths for importance sampling. The KL-divergence penalty coefficient $\beta$ is set to $0.001$, and the sampling temperature is 1.0. 

\paragraph{Reward Function Details.}
As implemented in our plugin, the total reward $R$ is a weighted sum of four specific components designed to enforce structure, accuracy, and grounding:
\begin{itemize}
    \item \textbf{Classification Reward ($R_{\text{clas}}$):} A dominant reward to ensure decision correctness. We assign $+4.0$ for correct predictions and $-4.0$ for incorrect ones.
    \item \textbf{Formatting Reward ($R_{\text{form}}$):} Capped at $1.0$, this component rewards the presence of mandatory XML tags (e.g., \texttt{<think>}, \texttt{<evidence>}) and valid bounding box syntax (e.g., \texttt{<|box\_start|>}).
    \item \textbf{Formatting Reward ($R_{\text{form}}$):} We also apply a strict format penalty of $-1.0$ if the detection verdict contradicts the localization output (e.g., predicting ``Real'' but generating bounding box coordinates, or predicting ``Forgery'' without coordinates).
    \item \textbf{Grounding Coverage Reward ($R_{\text{bbox}}$):} For correctly classified forgery samples, we reward the Mask Coverage $\mathcal{C}$ using a shaped sigmoid function:
    \begin{equation}
        R_{\text{bbox}} = \frac{0.25}{1 + e^{-10(\mathcal{C} - 0.5)}}
    \end{equation}
    This function scales the reward up to a maximum of $0.25$, effectively penalizing low overlaps while saturating for high coverage.
\end{itemize}

\end{document}